%% file: neurips_2024.tex
\documentclass{article}

\usepackage[preprint]{neurips_2024}

\usepackage[utf8]{inputenc} 
\usepackage[T1]{fontenc}    
\usepackage{hyperref}       
\usepackage{url}            
\usepackage{booktabs}       
\usepackage{amsfonts}       
\usepackage{nicefrac}       
\usepackage{microtype}      
\usepackage[table]{xcolor}         
\usepackage{algorithm}
\usepackage{stfloats}
\usepackage{algorithmic}
\usepackage{amsthm}
\usepackage{graphicx} 
\usepackage{anyfontsize}
\usepackage{tabularx}
\usepackage{multirow}
\usepackage{makecell}
\usepackage{stfloats}
\newtheorem{theorem}{Theorem}
\makeatletter

\newcommand{\ssymbol}[1]{^{\@fnsymbol{#1}}}
\input{math_commands}

\title{Federated Distillation for Medical Image Classification: Towards Trustworthy Computer-Aided Diagnosis}

\author{%
  Sufen Ren$\ssymbol{2}$, Yule Hu$\ssymbol{3}$, Shengchao Chen$\ssymbol{4}$, and Guanjun Wang$\ssymbol{8}$\\
  $\ssymbol{2}$ School of Information and Communication Engineering, Hainan University, China\\
  $\ssymbol{3}$ College of Mathematical Sciences, BoHai University, China \\
  $\ssymbol{4}$ Australian AI Institute, University of Technology Sydney, Australia\\
  $\ssymbol{8}$ School of Electronic Science and Technology, Hainan University, China \\
  Corresponding to: \texttt{shengchao.chen.uts@gmail.com, wangguanjun@hainanu.edu.cn}
}

\begin{document}

\maketitle

\begin{abstract}
Medical image classification plays a crucial role in computer-aided clinical diagnosis. While deep learning techniques have significantly enhanced efficiency and reduced costs, the privacy-sensitive nature of medical imaging data complicates centralized storage and model training. Furthermore, low-resource healthcare organizations face challenges related to communication overhead and efficiency due to increasing data and model scales. This paper proposes a novel privacy-preserving medical image classification framework based on federated learning to address these issues, named \textsc{FedMIC}. The framework enables healthcare organizations to learn from both global and local knowledge, enhancing local representation of private data despite statistical heterogeneity. It provides customized models for organizations with diverse data distributions while minimizing communication overhead and improving efficiency without compromising performance. Our \textsc{FedMIC} enhances robustness and practical applicability under resource-constrained conditions. We demonstrate \textsc{FedMIC}'s effectiveness using four public medical image datasets for classical medical image classification tasks. \footnotetext{Sufen Ren and Yule Hu contributed equally to this work.}
\end{abstract}

\section{Introduction}
Deep learning (DL)-based medical image classification (MIC) leverages computer vision techniques to automate the analysis and classification of extensive medical image dataset~\citep{chen2023interpretable}. Unlike traditional methods that depend on medical experts for multiple patient assessments~\citep{shi2023recognition}, this technology offers significant advantages in disease prevention, diagnosis, treatment planning, and patient management, enhancing both the accuracy and efficiency of diagnoses~\citep{li2023review}. Currently, DL models rely on extensive labeled datasets to achieve high accuracy through increased model complexity. However, traditional centralized training approaches, which process all data on a central server, conflict with the practicalities of medical care, particularly concerning patient privacy and data security~\citep{jiang2023fair}. Specifically, the transfer of patient data among healthcare providers is restricted by privacy and ethical regulations, limiting its use in centralized training. This restriction presents a significant challenge in developing efficient automated MIC models that can provide reliable support for clinical practice.

Existing related works based on advanced DL techniques mainly focus on improving through well-designed modules and learning strategies to optimize feature extraction capabilities~\citep{chen2023interpretable,wang2024ensemble,li2023review}. However, analyzing medical data using these methods via centralized cloud computing presents significant challenges in real-world applications. Specifically, these challenges include network dependency and privacy concerns:

\begin{itemize}
    \item[\textbf{(i)}] \textbf{Network Dependency:} Healthcare organizations generate vast amounts of patient data daily. Transmitting this data to a central server imposes a substantial communication burden, which is impractical for healthcare organizations with limited resources.
    \item[\textbf{(ii)}] \textbf{Privacy Concerns:} Electronic medical records of different patients are often distributed across various healthcare organizations. Sharing patient data between organizations is typically subject to stringent medical privacy protection and ethical regulations.
\end{itemize}
These constraints make it difficult for organizations to access a full range of medical images, limiting their ability to conduct effective analysis and obtain reliable insights. To address these issues, on-device intelligence for analyzing medical image data directly on the devices is crucial, as it reduces the need for data transfers, protects privacy, and decreases reliance on networks.

Federated Learning (FL)~\citep{mcmahan2017communication} is a promising paradigm for on-device intelligence that trains a uniform DL model collaboratively across multiple devices without revealing data. It is increasingly popular in  healthcare~\citep{dasaradharami2023comprehensive,li2023review}, recommendations~\citep{imran2023refrs,zhang2023comprehensive,zhang2023lightfr}, weather~\citep{chen2023prompt,chen2023spatial}, remote sensing~\citep{zhai2023fedleo,zhang2023federated}, and so on~\citep{chen2023collaborative,ren2024distributed}. Models trained by vanilla FL often underperform due to statistical heterogeneity, characterized by non-independent and identically distributed (Non.IID) data across devices. This challenge is exacerbated in federated MIC due to subtle inter-category differences and skewed data distribution among categories. These factors impede the training of a globally effective model. Developing strategies to enhance performance in the presence of significant statistical heterogeneity remains an open research challenge.

Shallow models often fail to capture the complex representations necessary for discerning key patient characteristics in medical images, including anatomical (shape and location of organs), pathological (masses and abnormal tissues), textural (grayscale distributions and localized texture patterns), and morphological features (edge sharpness and shape). While existing literature suggests adopting more complex structures to enhance model capabilities, this approach incurs higher computational costs~\citep{huo2024hifuse,chen2023interpretable,han2024dm}. These sophisticated models pose significant challenges for FL systems due to the frequent exchange of model updates between servers and clients—a process that becomes increasingly demanding with model complexity. This heightened communication burden is particularly problematic for resource-constrained devices in healthcare organizations.

To address these challenges, we propose a novel Personalized Federated Distillation for Medical Image Classification, dubbed \textsc{FedMIC}. This approach targets cross-healthcare organization medical image classification tasks with distributed, heterogeneous data, facilitating collaborative model training across databases while allowing personalized models without raw data sharing. \textsc{FedMIC} employs a Dual Knowledge Distillation (\textbf{Dual-KD}) strategy for local updating, enabling clients to develop custom local models that learn from both global and local knowledge without compromising privacy. Additionally, we introduce a Global Parameter Decomposition (\textbf{GPD}) method that compresses the full local inference model into low-rank matrices. By applying information entropy-based constraints, GPD filters out unnecessary parameters, significantly reducing communication costs while maintaining performance. This approach minimizes parameter transmission, making FedMIC suitable for low-resource environments.

We quantitatively evaluate the performance of the proposed \textsc{FedMIC} and typical FL algorithms using four publicly available medical image classification datasets, including both 2D and 3D images. The main contributions of this work are:
\begin{itemize}
    \item A privacy-preserving framework, \textsc{FedMIC}, for cross-healthcare organization medical image classification tasks.  This is the first solution to addresses the weak performance caused by the notorious heterogeneity while ensuring low communication costs.
    
    \item An effective dual knowledge distillation strategy within \textsc{FedMIC}. This method enables each client's student model to learn from both global and local knowledge while maintaining privacy, resulting in highly personalized models tailored to specific data distributions.
    
    \item A global parameter decomposition strategy during the aggregation process. This approach reduces communication overhead between clients and servers, improving efficiency by significantly decreasing transmission parameters while maintaining performance.
    
    \item  Extensive experiments on four publicly available real-world medical image classification datasets, demonstrating that \textsc{FedMIC} outperforms state-of-the-art FL algorithms and provides an efficient distributed learning strategy for low-resource scenarios.
\end{itemize}

\section{Related Work}
This section presents work related to our \textsc{FedMIC}, including Deep Learning-based Medical Image Classification (\textbf{Section.~\ref{subsec:dlmic}}) and Privacy-Preserving Federated Learning (\textbf{Section.~\ref{subsec:fl}}).
\subsection{Deep Learning-based Medical Image Classification}
\label{subsec:dlmic}
DL has substantially reduced automation costs across diverse applications~\citep{chen2023foundation,chen2023mask,chen2023tempee,peng2023diffusion,chen2022cmt,chen2022dynamic}. In medical image classification, DL techniques provide a more reliable and cost-effective solution for clinical analysis support, eliminating the need for high labor expenditures. However, medical image classification presents unique challenges compared to the analysis of natural images, particularly in capturing complex representations such as anatomical, textural, and morphological features.

While deep networks have been used to extract features from large-scale medical image datasets, effectively obtaining useful representations is still difficult. Recent efforts have focused on designing sophisticated modules for improved performance. For instance, Chen et al.~\citep{chen2023interpretable} introduced a hierarchical attention mechanism that enhances the performance of Transformers in multi-label classification while reducing spatio-temporal complexity. Huo et al.~\citep{huo2024hifuse} developed a three-branch hierarchical multi-scale feature fusion strategy that fuses global and local features without disrupting their modeling, thus improving efficiency. Han et al.~\citep{han2024dm} proposed a dynamic multi-scale convolutional neural network with an uncertainty quantification strategy to accelerate model convergence on medical images. Wang et al.~\citep{wang2024ensemble,wang2024less} suggested an ensemble learning strategy using multiple models pretrained on natural images to enhance the representation of complex medical images. Despite these advances, current models often overlook privacy concerns and are limited by their large parameter sizes, making them unsuitable for low-resource scenarios. Our study addresses these issues by focusing on a method that prioritizes privacy without compromising performance, particularly in low-resource environments.

\subsection{Privacy-Preserving Federated Learning}
\label{subsec:fl}
Federated Learning (FL)~\citep{mcmahan2017communication} is a  distributed machine learning paradigm that enables collaborative model training among multiple users without compromising data privacy. This approach has found applications in diverse fields, including recommendation systems~\citep{neumann2023privacy,zhang2023lightfr}, climate change~\citep{chen2023prompt,chen2023spatial,chen2024personalized}, and healthcare~\citep{rauniyar2023federated,zhang2023scoring}. However, FL faces two primary challenges: statistical heterogeneity (Non-IID) and communication efficiency. This paper provides a concise overview of recent advancements in addressing these challenges within the FL framework.

\paragraph{Statistical Heterogeneity in FL}  Various FL algorithms employ regularization techniques to ensure convergence in the presence of statistical heterogeneity. Li et al.~\citep{li2020federated} introduced a local regularization term to optimize individual client models, while Karimireddy et al. \citep{karimireddy2020scaffold} proposed a novel stochastic algorithm using control variates to address this challenge. Adaptations of Elastic Weight Consolidation (EWC) to FL scenarios have been explored by Shoham et al.~\citep{shoham2019overcoming} and Yao et al.~\cite{yao2020continual}. The former applied EWC to mitigate catastrophic forgetting during task transfer by selectively penalizing parameter vectors that deviate significantly from those learned in previous tasks. The latter utilized EWC to estimate the global model's importance weight matrix and incorporate each client's knowledge. Li et al.~\citep{li2021model} expanded on this concept by implementing regularization between global and local models, as well as between current and previous local models. To handle non-IID data and enhance training stability, Xu et al.~\citep{xu2022fedcorr} developed an adaptive weighted proximal regularization term based on estimated noise levels. While effective, these methods often incur high computational costs when dealing with large-scale parameter regularization.

Personalized Federated Learning (PFL) offers a more effective solution to the non-IID challenge than regularization-based methods by enabling the development of client-specific models. Several approaches have been proposed in this domain. pFedMe employs Moreau envelopes to create a regularized local objective, facilitating the separation of personalized model optimization from global learning~\citep{t2020personalized}. PerFedAvg, inspired by Model-Agnostic Meta-Learning, introduces a decentralized meta-learning framework to establish an initial model for rapid client adaptation~\citep{fallah2020personalized}. Although these approaches show promising results, most overlook the feature shift problem, with the notable exception of FedBN\citep{li2021fedbn}, which employs local batch normalization to mitigate this issue. Knowledge Distillation (KD)-based personalization techniques such as FedBE~\citep{chen2021fedbe} also help diminish the effects of non-IID by refining global and local representations. FedMD~\citep{li2019fedmd}, for instance, promotes inter-client learning by using knowledge distillation and transfer learning to align each local model with the global consensus. However, these strategies typically rely on averaging to form this consensus, heavily dependent on the quality of public data. To reduce this dependency, ensemble distillation using generated data and compressed federated distillation techniques~\citep{lin2020ensemble}, which incorporate distilled data curation, soft-label quantization, and delta-encoding, have been developed to lower communication costs~\citep{sattler2020communication}. Data-free knowledge distillation is another innovative approach where a server-derived lightweight generator~\citep{zhu2021data}, built solely from client model predictions, is distributed to clients. This generator uses the aggregated knowledge as an inductive bias to guide local training.

\paragraph{Communication Efficiency in FL} Heterogeneous environments can impair training efficiency. Consequently, enhancing communication efficiency and effectiveness has become a focal point in current efforts~\citep{chen2023prompt,chen2023spatial,bibikar2022federated}. Chen et al.~\citep{chen2023prompt,chen2023spatial} introduced a few parameters as prompts independent from the training model, and only activated and transmitted prompts during training and communication to reduce the computational burden. However, training a limited number of parameters often leads to a trade-off between performance and efficiency. Yao et al.~\citep{yao2019federated} reduces the number of communication rounds by integrating a maximum mean discrepancy constraint into the optimization objective. Dai et al.~\citep{dai2022dispfl} implements a decentralized sparse training approach, allowing each local model to utilize a personalized sparse mask to identify active parameters, which remain consistent during local training and peer-to-peer communication. This method requires only the initial transmission of the indices of active parameters, followed by the transmission of their values in subsequent communications, thereby substantially lowering communication costs. In this paper, we propose \textsc{FedMIC}, a parameter-efficient personalized method for federated medical image classification. \textsc{FedMIC} aims to provide highly customized models for each client, improving learning in the face of heterogeneity while significantly reducing communication overhead between clients and the server.

\section{Preliminaries}
\paragraph{Federated Learning} Under Vanilla FedAvg~\citep{mcmahan2017communication}, a central server coordinates $N$ clients to collaboratively train a uniform global model intended to generalize across all devices.  Specifically, in each communication round $t$, the server samples a fraction of all the clients, $C$, to join the training. The server distributes the global model $w$ to these clients. Each selected client received global model on their private dataset $D_k - P_k$ ($D_k$ obeys the distribution $P_k$) to obtain $w_k$ through local training process $w_k \leftarrow w - \eta \nabla \ell (w; x_i, y_i), (x_i, y_i) \in D_k$. The $k$-th client uploads the trained local model $w_k$ to the server that aggregates them to update the global model via $w = \sum_{k = 0}^{K-1} \frac{n_k}{n} w_k$. FedAvg aims to minimize the average loss of the global model $w$ on all clients' local datasets:
\begin{equation}
    F(w) \text{:}= \mathop{\argmin}\limits_{w_1, w_2, ..., w_N}  \sum_{k=1}^{N} \frac{n_k}{n} F_k(w_k)
\end{equation}
where $n_k$ is the number of samples stored by the $k$-th client. $n$ is the number of samples held by all clients. $F_k(w_k)$ denotes the local objective function of $k$-th client that can be formulated as $\gL_k(w_k) = \ell_k (w_k; (x_i, y_i))$, where the $\gL$ is loss function. 

\paragraph{Problem Formulation for FedMIC} The core framework of our proposed \textsc{FedMIC} aligns with the standard FedAvg protocol, with key distinctions in data distribution and resource constraints. In \textsc{FedMIC}, each client possesses a unique distribution of medical image data, creating a statistically heterogeneous environment. Moreover, healthcare organizations typically operate with limited computational resources, facing challenges in managing large-scale data and complex neural networks. This setup reflects real-world constraints in storage, communication, and processing capabilities. Consequently, \textsc{FedMIC} addresses two primary challenges:  (1) \textit{mitigating the impact of statistical heterogeneity on Federated Medical Image Classification} and (2) \textit{reducing communication overhead and enhancing efficiency without compromising model effectiveness.}

\section{Methodology}
This section elaborates on the details of our proposed framework, \textsc{FedMIC}. This section is divided into three parts, including \textbf{(1)} Framework Architecture of \textsc{FedMIC} (\textbf{Section.~\ref{subsec:arch}}), \textbf{(2)} Dual Knowledge Distillation (Dual-KD) (\textbf{Section.~\ref{subsec:lcd}}), and \textbf{(3)} Global Parameter Decomposition (GPD) (\textbf{Section.~\ref{subsec:gpd}}).

\subsection{Framework Architecture of \textsc{FedMIC}}
\label{subsec:arch}
\begin{figure}[tbh]
    \centering
    \includegraphics[width=1\textwidth]{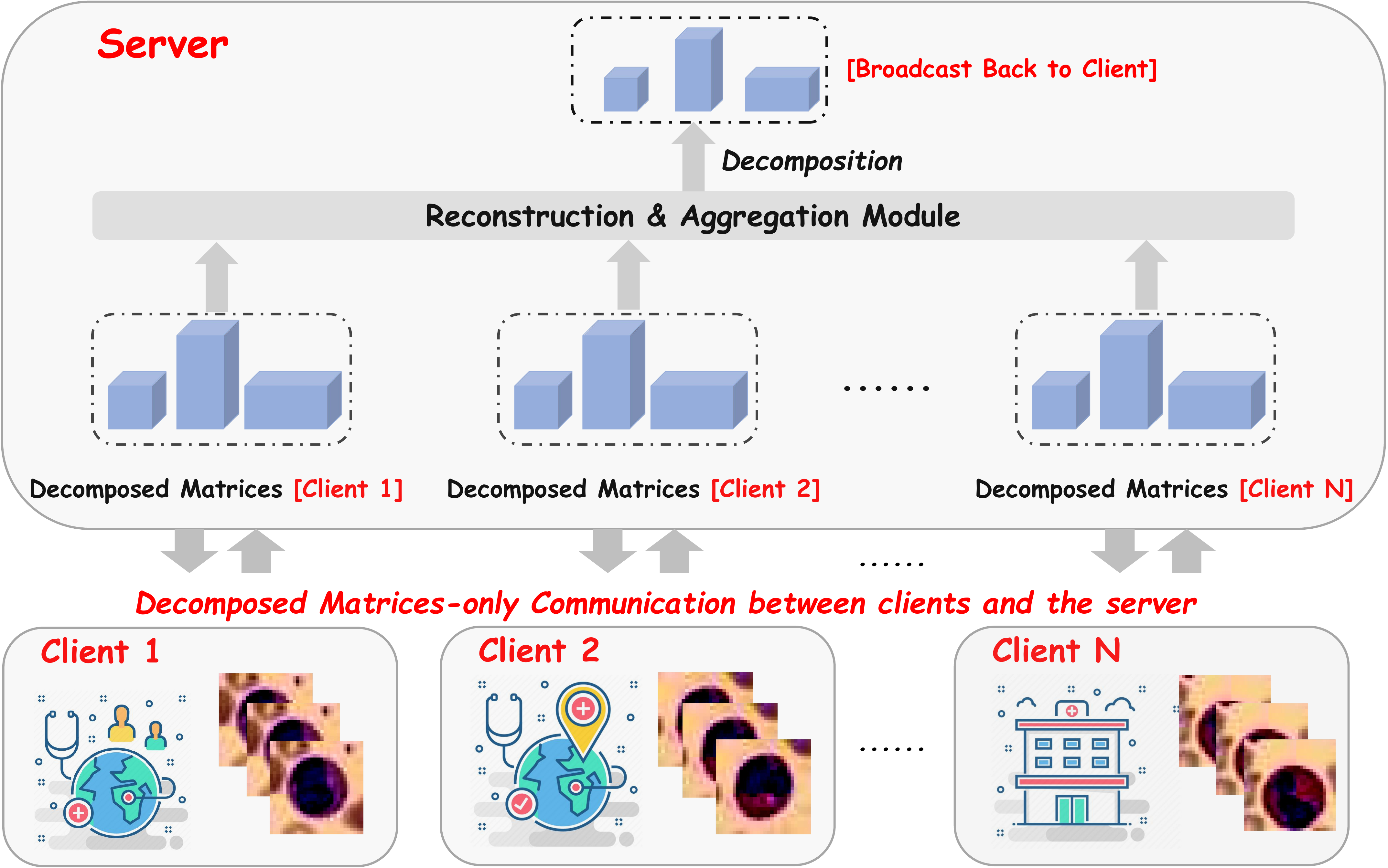}
    \caption{Schematic diagram of our \textsc{FedMIC}. Each healthcare organization as an independent client with private medical images that remain unshared throughout the training process. Clients train their local models using exclusively local data before transmitting parameters to the central server. FedMIC significantly reduces communication overhead by transmitting only a small subset of parameters from the decomposition matrix, rather than the entire local model. The server reconstructs and aggregates these uploaded parameters within its service area, subsequently broadcasting the aggregated parameters to all clients for the next iteration of training and communication.}
    \label{fig:framework}
\end{figure}

Architecture of our \textsc{FedMIC} is as shown in Figure~\ref{fig:framework}, which comprising $N$ clients denote different healthcare organization, and a central server. Each client maintains two local model, a local teacher model and a local student model, which is work during inference, while the teacher model does not work during inference. In addition, each client does not share any local data during training, thus ensuring the privacy of each health organization. The central server does not access any client's data.

In contrast to FedAvg~\citep{mcmahan2017communication}, \textsc{FedMIC} employs a dual-model approach where clients update both a local student model and a local teacher model during each global communication round. However, only the student model parameters are transmitted to the server, while the teacher model remains local, facilitating personalized knowledge distillation. The server aggregates the received student model parameters and disseminates the updated model to all clients for the subsequent communication round. This iterative process continues until the student model converges. Unlike traditional centralized training methods, this approach eliminates the need for raw data sharing, significantly reducing the transmission of sensitive information. Consequently, the framework enhances privacy protection for participating healthcare organizations during model training.

\begin{figure}[tbh]
    \centering
    \includegraphics[width=1\textwidth]{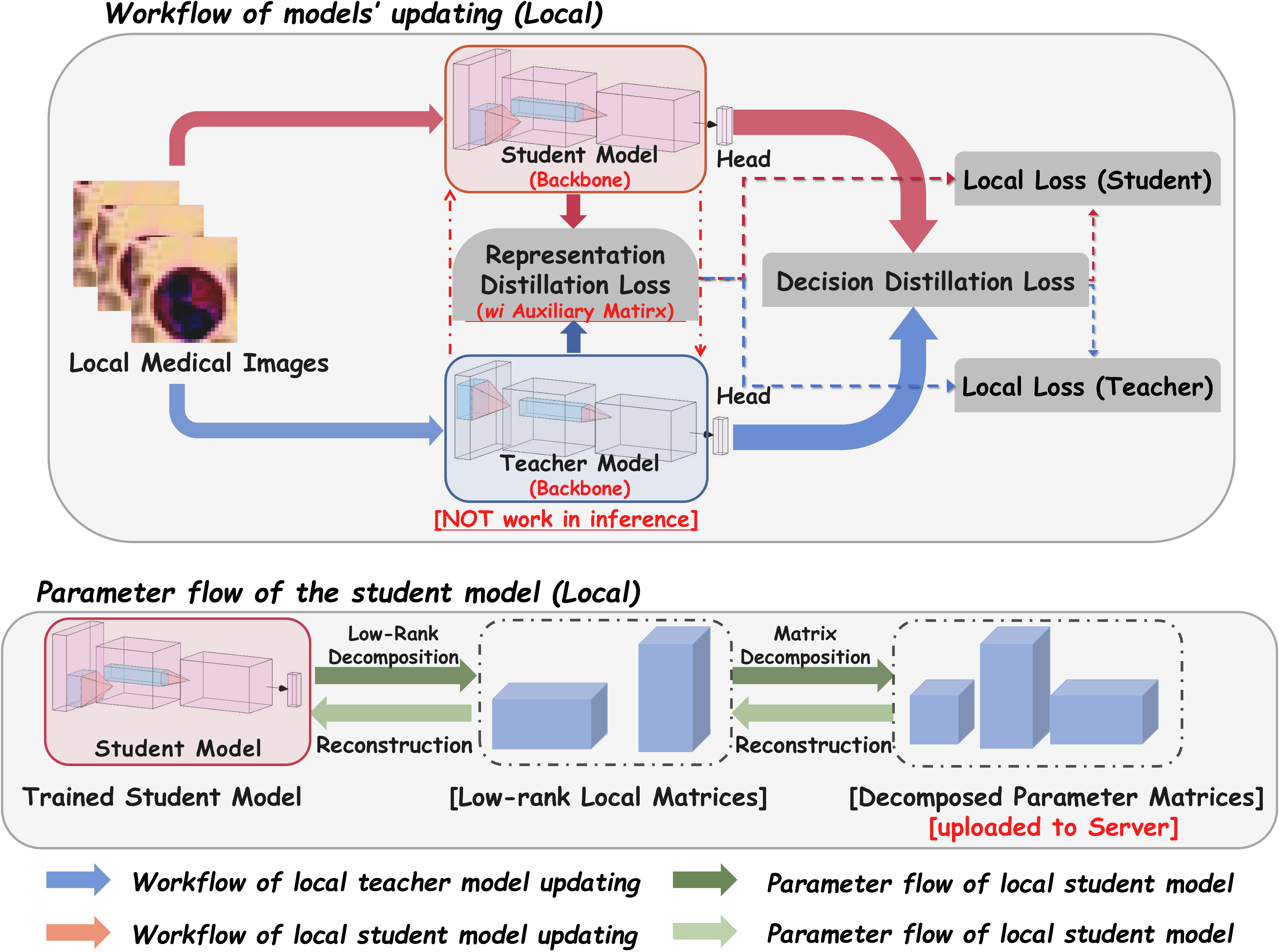}
    \caption{Schematic diagram of the process of local updating within our \textsc{FedMIC}. The local update comprises two primary phases: (1) local model updating and (2) parameter flow of the updated local model. In phase one, duplicate local medical image data are input into both student and teacher models. The extracted representations are used to compute the representational distillation loss, facilitating bi-directional correction. These representations are then fed into the corresponding model heads for decision-making, where the decision distillation loss is calculated using two independent auxiliary matrices. This process incorporates both the model's classification loss and the decision distillation loss between student and teacher models. In phase two, the trained student model's parameters undergo low-rank decomposition and matrix decomposition, resulting in a decomposed parameter matrix that is uploaded to the server. The reconstruction process involves the inverse processing of these parameters, which are subsequently broadcast from the server to the client.}
    \label{fig:local}
\end{figure}

\subsection{Dual Knowledge Distillation Mechanism on Local Updating}
\label{subsec:lcd}

Knowledge Distillation (KD) enables simpler models to learn from more complex ones, enhancing their performance through knowledge transfer~\citep{habib2023knowledge}. In real-world medical applications, data heterogeneity among healthcare organizations poses significant challenges, rendering global models ineffective across diverse data distributions. To address this Non-IID challenge and its negative impact on model performance, we propose a novel KD strategy within FedMIC, termed Dual Knowledge Distillation (Dual-KD). Unlike traditional KD approaches that focus on local parameter reduction or global-to-local model distillation, our Dual-KD encourages student models to integrate insights from both local and global knowledge, bridging the gap between local and global representations.

The local updating process within \textsc{FedMIC} is shown in Figure~\ref{fig:local}. Dual-KD employs two structurally consistent networks: a teacher model and a student model. During local updates, both models learn concurrently from local medical imaging data. Crucially, only the student model's parameters are uploaded to the server for global collaborative learning, enabling it to incorporate global knowledge. The teacher model, updated locally, enhances its capacity for personalized learning based on local data. Dual-KD facilitates mutual learning between the models; the student model benefits from acquired global insights, while the teacher model develops robust, personalized representations. This approach not only tailors local models to the specific needs of Non-IID clients but also facilitates global knowledge sharing without compromising privacy, thereby preventing data silos.

To enhance knowledge transfer between local teacher models and student models, which includes both locally personalized knowledge and insights aimed at global understanding, we introduce two distinct local optimization objectives within our Dual-KD strategy: Representation Distillation Loss (RDL, $\gL_{rep}$) and Decision Distillation Loss (DDL, $\gL_{dec}$). The former is used to align the potential representations from teacher and student models, and the latter is used correct each other between classificationdecisions on potential representations of medical images. For these objectives, we divide both the teacher and student models into two components: the $\texttt{Backbone}$, responsible for data representation extraction, and the $\texttt{Head}$, which acts as a classifier utilizing these representations to make decisions. The local medical images on client $i$ is labeled as $\mX_i$, with the predictions from the local teacher and student models denoted as $\mY^t_i$ and $\mY^s_i$, respectively. To counteract potential performance degradation due to misleading backbone representations, we introduce a trainable auxiliary matrix $\mW_{aux}$. This matrix ensures accurate updates to both models, facilitating bidirectional corrections. The process of extracting representations is described as follows:
\begin{equation}
    \mH^s_{hs} = \texttt{Backbone}^{s}(\mX_i),  \qquad \mH^t_{hs} = \texttt{Backbone}^{t}(\mX_i),
\end{equation}
where the $\texttt{Backbone}^{s}$ and $\texttt{Backbone}^{t}$ denote the backbone of the teacher and the student model, respectively. Then the process of dual-way correction (Representation Distillation Loss in Figure 2) can be formulated as:
\begin{equation}\label{rep_loss}
    \gL_{rep} = \frac{1}{n} \sum_{i=1}^{n}(\mH^s_{hs} \cdot \mW_{aux} - \mH^t_{hs} \cdot \mW_{aux})^2.
\end{equation}
To construct the above-mentioned Decision Distillation Loss, we use the target label to compute the task losses for both the teacher model and the student model. Task losses for these two models can be formulated based on standard Cross-Entropy Loss:
\begin{equation}
    \gL^t_{task} = -\sum_{i} \mY^t_i \log(\mY), \qquad \gL^s_{task} = -\sum_{i} \mY^s_i \log(\mY),
\label{task:loss}
\end{equation}
where the $\mY$ is the distribution of ground truth of local label, $\mY^t_i$ and $\mY^s_i$ can be expressed as $\texttt{Head}^s(\mH^s_{hs})$ and $\texttt{Head}^t(\mH^s_{hs})$ according to the above-mentioned \texttt{Backbone-Head} setting. The DDL consists of two part: representation-based part and decision-based part, for achieving comprehensive distillation among student and teacher models. The representation-based part of DDL ($\gL_{dec-r}$) for both teacher and student models are formulated below:
\begin{equation}
    \gL^s_{dec-r} = \gL^t_{dec} = \frac{\gL_{rep}}{\gL^t_{task} + \gL^s_{task}}.
\label{eq:dec_1}
\end{equation}
In this way, the distillation intensity is weak if the prediction of teacher and student models is incorrect (task losses $\gL^t_{task}$ and $\gL^s_{task}$ are large). The distillation is highly effective when the teacher and student models are well trained (task losses $\gL^t_{task}$ and $\gL^s_{task}$ is small), which have the ability to avoid over-fitting and enforcing the extracted latent representation from backbone both teacher and student models toward balance and alignment (have similar knowledge). 

In addition, we further introduce decision-based DDL ($\gL_{dec-d}$) that is utilized to distillate the teacher model and student model directly based on their output prediction according to Eq.(\ref{task:loss}), as follows:
\begin{equation}
    \gL^t_{dec-d} = \frac{-\sum_i \mY^t \log (\frac{\mY^s}{\mY^t})}{\gL^t_{task} + \gL^s_{task}}, \qquad  \gL^t_{dec-d} = \frac{-\sum_i \mY^s \log (\frac{\mY^t}{\mY^s})}{\gL^t_{task} + \gL^s_{task}},
\label{eq:dec_2}
\end{equation}
where the term of $-\sum_i \mY^t \log (\frac{\mY^s}{\mY^t})$ and $-\sum_i \mY^s \log (\frac{\mY^t}{\mY^s})$ is the Kullback–Leibler divergence (KLD) often appears in conventional knowledge distillation tasks.

Based on the above-proposed loss function, We can summarize the local loss for both teacher model (denotes $\gL^t$) and student model (denotes $\gL^s$) local optimization as follows:
\begin{equation}\label{local:loss}
    \gL^t = \underbrace{\gL^t_{dec-d} + \gL^t_{dec-r}}_{DDL} + \gL^t_{task}, \qquad \gL^s_{dec} = \underbrace{\gL^s_{dec-d} + \gL^s_{dec-r}}_{DDL} + \gL^s_{task},
\end{equation}
Both the teacher and student models are locally updated using the same optimizers on each client. The student model's gradients $\rvg$ on the $i$-th client are calculated as $\rvg^s_i = \frac{\partial \gL^s_{bik}}{\partial \Theta^s}$, where $\Theta^s$ represents the student model's parameters. Similarly, the teacher model's gradients on the same client are obtained using $\rvg^t_i = \frac{\partial \gL^t_{bik}}{\partial \Theta^t}$, with $\Theta^t$ being the teacher model's parameters.

Once both models are updated, each client transmits student model's parameters $\Theta^s$ to the central server. The server employs the FedAvg algorithm~\citep{mcmahan2017communication} to aggregate these parameters into a global model, which it then redistributes to the clients. Subsequently, clients update their student models with the globally aggregated parameters. Concurrently, the local teacher model continues to offer personalized instruction and knowledge transfer. This process repeats until convergence is achieved for both the student and teacher models.

\subsection{Global Parameter Decomposition on Global Aggregation}
\label{subsec:gpd}
The communication overhead in FL frameworks is largely determined by the volume of parameters transmitted from clients to the server. This overhead poses a significant challenge for systems comprising low-resource devices with limited computational power and communication bandwidth. In resource-constrained environments, increased communication costs and reduced efficiency may compromise the accuracy of decision-making systems. To address this issue, we introduce Global Parameter Decomposition (GPD), a simple yet effective method to compress parameters exchanged during server-client communications. Inspired by low-rank parameter matrix decomposition (LoRA)~\citep{hu2021lora}, GPD aims to reduce communication costs without sacrificing performance. The locally updated student model parameters are decomposed into two parts:
\begin{equation}
    \rvg^s_i = \rvg^{s,p}_i \cdot \rvg^{s, n}_i, \quad
    \textit{where} \quad \rvg^s_i \in \sR^{P\times Q}, \rvg^{s, p}_i \in \sR^{P\times r}, \rvg^{s, n}_i \in \sR^{r\times Q}.
\end{equation}
For simplicity, we use a two-dimensional representation where the original student model parameters are split into two low-rank matrices, $\rvg^{s, p}_i$ and $\rvg^{s, n}_i$, with rank $r$. We define $r$ as the ratio $Q // P$ (if $Q > P$) or $P // Q$ (if $P > Q$). Prior to uploading, local gradients are decomposed into smaller matrices using singular value decomposition (SVD). The server reconstructs local parameters by multiplying these matrices before aggregation. The aggregated global parameters are then decomposed and sent to clients for reconstruction during model updates. Specifically, we apply SVD to the matrices $\rvg^{s,p}_i$ and $\rvg^{s,n}_i$:
\begin{equation}
    \rvg^{s, p}_i \approx \emU^p \Sigma^p \emV^{p}, \quad \rvg^{s, n}_i \approx \emU^n \Sigma^n \emV^{n}
\end{equation}
where $\emU^p \in \sR^{Q \times K}$, $\Sigma^p \in \sR^{K \times K}$, $\emV^p \in \sR^{K \times r}$, $\emU^n \in \sR^{r \times K}$, $\Sigma^n \in \sR^{K \times K}$, $\emV^n \in \sR^{K \times Q}$, and $K$ is the number of retained singular values. In terms of the total number of parameters, if the value of $K$ meets $QK + K^2 + Kr < Pr $ (for $\rvg^{s, p}$), or $rK + K^2 + KQ < Pr $ (for $\rvg^{s, n}$), the uploaded parameter from each client and downloaded parameter from the server can be reduced significantly. In multi-dimensional CNNs utilized in MIC image classification, different parameter matrices (e.g., convolution units, linear layers, etc.) are decomposed independently, and the global parameters on the server are decomposed in the same way. We denote the singular values of $\rvg^{s, p}_i$ as $[\sigma_1,\sigma_2,\cdots,\sigma_r]$, $\rvg^{s, n}_i$ as $[\sigma_1,\sigma_2,\cdots,\sigma_Q]$ ordered by their absolute values. To minimize approximation error, we employ the Variance Explained Criterion~\citep{henseler2015new} to dynamically adjust the number of transmitted singular values, optimizing benefits and minimizing communication overhead in resource-limited environments, as follows:
\begin{equation}
    \min_K \frac{\sum_{i=1}^K \sigma_i^2}{\sum_{i=1}^Q \sigma_i^2} \textgreater \alpha.
\end{equation}
This strategy enables clients to transmit parameters with maximum relevant information, preventing the negative effects of unnecessary parameter transmission on model performance and communication costs. The dynamic selection of parameters based on their performance on private client data adds flexibility to the approach. The algorithm implement of \textsc{FedMIC} is shown in Alg.~\ref{alg1} and Alg.~\ref{alg2}.

\begin{algorithm}[tb]
   \caption{Implement of the proposed \textsc{FedMIC}}
    \begin{algorithmic}
   \STATE Setting the local learning rate $\eta$, client number $N$, and local dataset $D_i=\{(x_1,y_1), (x_2,y_2), ..., (x_n,y_n)\}$
   \STATE Setting communication rounds $R$, local updating steps $e$
   \STATE Setting hyperparameters $\alpha$,
   \FOR{communication rounds $R=1,2,3...$}
   \STATE \colorbox{green! 40}{\bf Client-side:}
   \FOR{each client $i$ in parallel}
   \STATE Initialize student model $\Theta^s_i$ and teacher model $\Theta^t_i$
   \STATE $\Theta^t_i \leftarrow \textsc{LocalUpdate}(D_i, \eta, \Theta^t_i/ \rvg^t_i, e)$ 
   \STATE $\Theta^s_i \leftarrow \textsc{LocalUpdate}(D_i, \eta, \Theta^s_i, e)$ 
   \STATE $\rvg^{s,p}_i, \rvg^{s,n}_i  \leftarrow \textsc{Decom}(\rvg^s_i, e)$
   \STATE $\emU^p \Sigma^p \emV^{p} \leftarrow \textsc{GPD}(\rvg^{s,p}_i, e)$
   \STATE $\emU^n \Sigma^n \emV^{n} \leftarrow \textsc{GPD}(\rvg^{s,n}_i, e)$
   \STATE Clients upload $\emU^p_i \Sigma^p_i \emV^{p}_i$ and $\emU^n_i \Sigma^n_i \emV^{n}_i$ to the server
   \ENDFOR
   \STATE \colorbox{red! 40}{\bf Server-side:}
   \STATE Server receives $\emU^p \Sigma^p \emV^{p}$ and $\emU^n \Sigma^n \emV^n$
   \STATE Server reconstructs $\rvg^t_i$
   \STATE Server aggregates $\rvg \leftarrow \sum_{i}^{N} \frac{n_i}{n}\rvg^t_i$
   \STATE $\rvg \leftarrow \emU \Sigma \emV$
   \STATE Server distribute $\rvg \leftarrow \emU \Sigma \emV$ to each client
   \ENDFOR
\end{algorithmic}
\label{alg1}
\end{algorithm}

\begin{algorithm}[tb]
   \caption{Implement of \textsc{LocalUpdate}}
    \begin{algorithmic}
    \STATE Setting the local student parameter $\Theta^s$, the teacher model parameter $\Theta^t$, and local dataset $D$.
   \STATE \colorbox{black! 40}{\bf LocalUpdate$(D, \eta, \Theta, e)$:}
   \FOR{local updating steps $e=1,2,3...$}
   \STATE Local model $\Theta^s$ and $\Theta^t$ trained by private data $D$
   \STATE Compute task losses according to Eq.~\ref{task:loss}
   \STATE Compute representation distillation losses according to Eq.~\ref{rep_loss}
   \STATE Compute representation-based decision distillation losses according to Eq.~\ref{eq:dec_1}
   \STATE Compute decision-based decision distillation losses according to Eq.~\ref{eq:dec_2}
   \STATE Compute local losses for both student and teacher models according to Eq.~\ref{local:loss}
   \STATE Parameter update $\Theta^s (new) \leftarrow  \Theta^s (old)$, $\Theta^t (new) \leftarrow  \Theta^t (old)$
   \ENDFOR
\end{algorithmic}
\label{alg2}
\end{algorithm}

\section{Theorems and Proofs}
In this work, we actually consider the following distributed optimization model:
\begin{equation}
    \min_w \left\{F(w) = \sum^N_{k=1}p_k F_k(w)\right\}
\end{equation}
where $N$ is the number of devices, and $p_k$ is the weight of the $k$-th device such that $p_k \geq 0$ and $\sum_{k=1}^N p_k = 1$. Suppose the $k$-th device holds the $n_k$ training data: $x_{k,1}, x_{k,2}, ..., x_{k,n_k}$. The local objective $F_k(\cdot)$ is defined by
\begin{equation}
\begin{aligned}
     &F_k(w) = \frac{1}{n_k}\sum^{n_k}_{j=1} \gL(w;x_{k,j}), \\
     & \textit{where} \quad \gL(w;x_{k,j})  := \left \{\gL^t,\gL^s \right\}\\
    & := \left \{\gL^t_{dec-d} + \gL^t_{dec-r} + \gL^t_{task}, \gL^s_{dec-d} + \gL^s_{dec-r} + \gL^s_{task} \right\}
\end{aligned}
\end{equation}
Considering that the student and teacher models are architecturally consistent and updated in parallel during the local update of \textsc{FedMIC}, we will subsequently consider only the student model's in order to simplify the relevant theorems and proofs, which has no impact on the results.

\begin{theorem}[Generalization Bound of \textsc{FedMIC}]
Consider a on-device medical image classification system with $m$ clients (devices). Let $\gD_1, \gD_2, ..., \gD_m$ be the true data distribution and $\hat{\gD_1}, \hat{\gD_2}, ... , \hat{\gD_m}$ be the empirical data distribution. Denote the head $h$ as the hypothesis from $\gH$ and $d$ be the VC-dimension of $\gH$. The total number of samples over all clients is $N$. Then with probability at least $1-\delta$:
\begin{equation}\small
    \begin{aligned}
        &\max_{(\lbrace \mP_1 \rbrace, \lbrace \mP_2 \rbrace, ..., \lbrace \mP_m \rbrace)} \left| \sum_{i=1}^m \frac{|D_i|}{N}\gL_{\gD_i}(\theta_i;\cdot) - \sum_{i=1}^m \frac{|D_i|}{N} \gL_{\hat{\gD_i}}(\theta_i;\cdot) \right| \\
        &\leq \sqrt{\frac{N}{2} \log\frac{(m+1) |\mP|}{\delta}} + \sqrt{\frac{d}{N}\log\frac{eN}{d}}.
    \end{aligned}
\end{equation}
\end{theorem}

\textit{Proof.}  We start from the McDiarmid's inequality as
\begin{equation}
    \sP[g(X_1, ..., X_n) - \mathbb{E}[g(X_1, ..., X_n)] \geq \epsilon] \leq \exp{(-\frac{2\epsilon^2}{\sum_{i=1}^n c_i^2})}
\end{equation}
when
\begin{equation}
    \sup_{x_1, ..., x_n} |g(x_1,x_2,...,x_n) - g(x_1, x_2,...,x_n)| \leq c_i
\end{equation}
Eq.~15 equals to
\begin{equation}
    \sP[g(\cdot) - \mathbb{E}[g(\cdot)]\leq \epsilon] \geq 1 - \exp{(-\frac{2\epsilon^2}{\sum_{i=1}^n c_i^2})}
\end{equation}
which means that with probability at least $1 - \exp{(-\frac{2\epsilon^2}{\sum_{i=1}^n c_i^2})} $,
\begin{equation}
    g(\cdot) - \mathbb{E}[g(\cdot)] \leq \epsilon
\end{equation}
Let $\delta = \exp{(-\frac{2\epsilon^2}{\sum_{i=1}^n c_i^2})}$, the above can be rewritten as with the adaptive prompts at least $1 - \delta$,
\begin{equation}
    g(\cdot) - \mathbb{E}[g(\cdot)] \leq \sqrt{\frac{\sum_{i=1}^n c_i^2}{2} \log\frac{1}{\delta}}
\end{equation}
Now we substitute $g(\cdot)$:
\begin{equation}
    \max_{(\theta_1, \theta_2,...,\theta_m)} \left( \sum_{i=1}^m \frac{|D_i|}{N}\gL_{\gD_i}(\theta_i;\cdot) - \sum_{i=1}^m \frac{|D_i|}{N} \gL_{\hat{\gD_i}}(\theta_i;\cdot) \right)
\end{equation}
we can obtain that with probability at least $1-\delta$, the following holds for specific adaptive prompts,
\begin{equation}\small
    \begin{aligned}
        & \max_{(\theta_1, \theta_2,...,\theta_m)} \left( \sum_{i=1}^m \frac{|D_i|}{N}\gL_{ \gD_i}(\theta_i;\cdot) - \sum_{i=1}^m \frac{|D_i|}{N} \gL_{\hat{\gD_i}}(\theta_i;\cdot) \right)\\
        & - \mathbb{E} \left[\max_{(\theta_1, \theta_2,...,\theta_m)} \left( \sum_{i=1}^m \frac{|D_i|}{N}\gL_{\gD_i}(\theta_i;\cdot) - \sum_{i=1}^m \frac{|D_i|}{N} \gL_{\hat{\gD_i}}(\theta_i;\cdot)\right) \right] \\ 
        & \leq \sqrt{\frac{N}{2} \log \frac{1}{\delta}}
    \end{aligned}
\end{equation}
Considering there are $(m+1) |\mP|$ singular value in total, by using Boole's inequality, with probability at least $1-\delta$, the following holds,
\begin{equation}\small
    \begin{aligned}
        & \max_{(\theta_1, \theta_2,...,\theta_m)} \left( \sum_{i=1}^m \frac{|D_i|}{N}\gL_{ \gD_i}(\theta_i;\cdot) - \sum_{i=1}^m \frac{|D_i|}{N} \gL_{\hat{\gD_i}}(\theta_i;\cdot) \right) \\
        & \leq \mathbb{E} \left[\max_{(\theta_1, \theta_2,...,\theta_m)} \left( \sum_{i=1}^m \frac{|D_i|}{N}\gL_{\gD_i}(\theta_i;\cdot) - \sum_{i=1}^m \frac{|D_i|}{N} \gL_{ \hat{\gD_i}}(\theta_i;\cdot) \right) \right] \\
        &+  \sqrt{\frac{N}{2} \log \frac{(m+1) |\mP|}{\delta}}
    \end{aligned}
\end{equation}
where $N$ is the total number of samples over all clients.
\begin{equation}
    \begin{aligned}
        &\mathbb{E} \left[\max_{(\theta_1, \theta_2,...,\theta_m)} \left( \sum_{i=1}^m \frac{|D_i|}{N}\gL_{\gD_i}(\theta_i;\cdot) - \sum_{i=1}^m \frac{|D_i|}{N} \gL_{ \hat{\gD_i}}(\theta_i;\cdot) \right) \right] \\
        &\leq \mathbb{E} \left[\sum_{i=1}^m \frac{|D_i|}{N} \max_{\lbrace \mP_i \rbrace}   \left(\gL_{\gD_i}(\theta_i;\cdot) - \gL_{\hat{\gD_i}}(\theta_i;\cdot) \right) \right] \\
        & \leq^{a} \sum_{i=1}^m \frac{|D_i|}{N} \mathcal{R} (\gH) \\
        & \leq \sum_{i=1}^m \frac{|D_i|}{N} \sqrt{\frac{d}{|D_i|} \log \frac{e|D_i|}{d}} \\
        & \leq \sum_{i=1}^m \frac{D_i}{N} \sqrt{\frac{d}{|D_i|} \log \frac{eN}{d}} \\
        &\leq^b \sqrt{\frac{d}{N}\log \frac{eN}{d}}
    \end{aligned}
\end{equation}
where $\gH$ is the hypothesis set of head $h$, $d$ is the VC-dimension of $\gH$. The $a$ follow from the definition of Rademacher complexity
\begin{equation}
    \mathcal{R}_n(\mathcal{F}) = \mathbb{E}_{\sigma}\left[\sup_{f\in\mathcal{F}}\frac{1}{n}\sum_{i=1}^{n}\sigma_i f(x_i)\right],
\end{equation}
where $\sigma_1, \sigma_2, \ldots, \sigma_n$ are independent Rademacher random variables that take values in $\lbrace -1, 1 \rbrace$ with equal probability,
$\mathbb{E}_{\sigma}$ denotes the expectation over the Rademacher variables,
$x_1, x_2, \ldots, x_n$ are the input data points, and the $b$ follows from Jensen's inequality, so
\begin{equation}
    \begin{aligned}
        & \max_{(\theta_1, \theta_2,...,\theta_m)} \left| \sum_{i=1}^m \frac{|D_i|}{N}\gL_{ \gD_i}(\theta_i;\cdot) - \sum_{i=1}^m \frac{|D_i|}{N} \gL_{\hat{\gD_i}}(\theta_i;\cdot) \right| \\
        & \leq \sqrt{\frac{N}{2} \log\frac{(m+1) |\mP|}{\delta}} + \sqrt{\frac{d}{N}\log\frac{eN}{d}}
    \end{aligned}
\end{equation}

\section{Experiments and Results}
In this section, we describe the dataset and the experimental setup. We then present the results conducted in Non-IID settings. These include standard experiment reports, ablation studies on key components, assessments of parameter impacts, and experiments on differential privacy.

\subsection{Dataset and Pre-Processing}
Our experiments utilized four representative Medical Image Classification (MIC) datasets derived from the publicly available MedMNIST: BloodMNIST, TissueMNIST, OrganMNIST(2D), and OrganMNIST(3D). These datasets are publicly available at \url{https://medmnist.com/}. All images were resized to a standard $64 \times 64$ resolution. To simulate the data distribution across different clients in federated learning, we partitioned the images into 20 distinct subsets. For each subset (client), data was divided into training, testing, and validation sets in a 7:2:1 ratio.Specific information about the datasets such as the number of samples, some sample examples can be found in \textbf{Table~\ref{tab:dataset_exm}}.

\begin{table}[tbh]
    \centering
    \caption{Detailed information of the datasets used as well as schematic diagrams including BloodMNIST, TissueMNIST, OrganMNIST2D and OrganMNIST3D. Under FL framework, we assigned them to 20, 20, 20, and 10 clients to simulate Non-IID in real-world applications, respectively.}
    \resizebox{\textwidth}{!}{
    \begin{tabular}{>{\centering\arraybackslash}m{0.2\textwidth}|>{\centering\arraybackslash}m{0.15\textwidth}|>{\centering\arraybackslash}m{0.15\textwidth}|>
    {\centering\arraybackslash}m{0.1\textwidth}|>{\centering\arraybackslash}m{0.4\textwidth}}
        \toprule
        \textbf{Dataset} & \textbf{Num. of Class} & \textbf{Num. of Sample} & \textbf{Num. of Client} & \textbf{Examples (four randomly sampled)} \\
        \midrule
        \makecell{BloodMNIST}  & 8 & 17,092 & 20 & \includegraphics[width=0.4\textwidth]{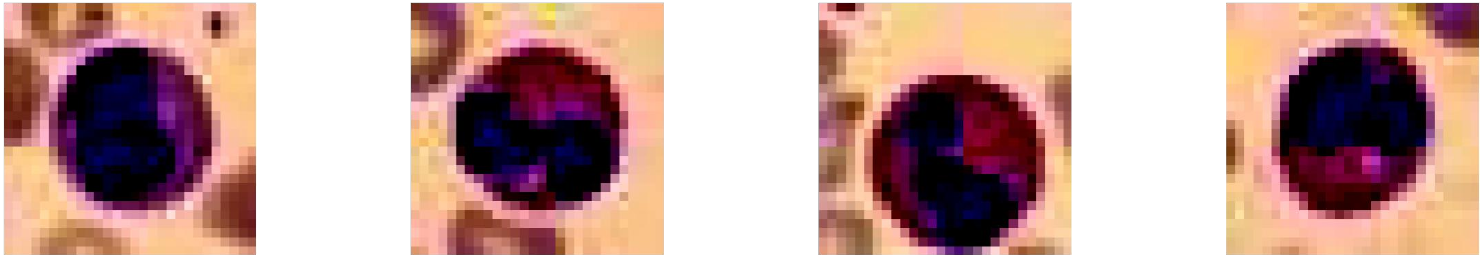} \\
        \midrule
        \makecell{TissueMNIST} & 8  & 236,386 & 20 & \includegraphics[width=0.4\textwidth]{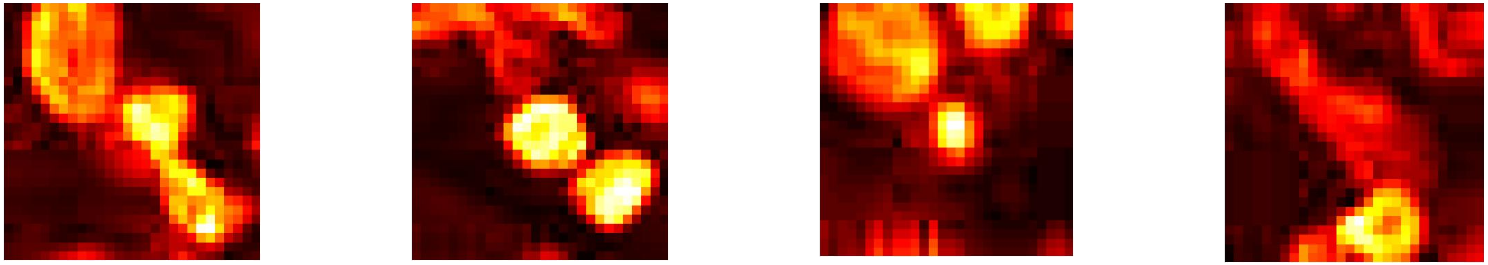} \\
        \midrule
        \makecell{OrganMNIST (2D)} & 11  & 25,211 & 20 & \includegraphics[width=0.4\textwidth]{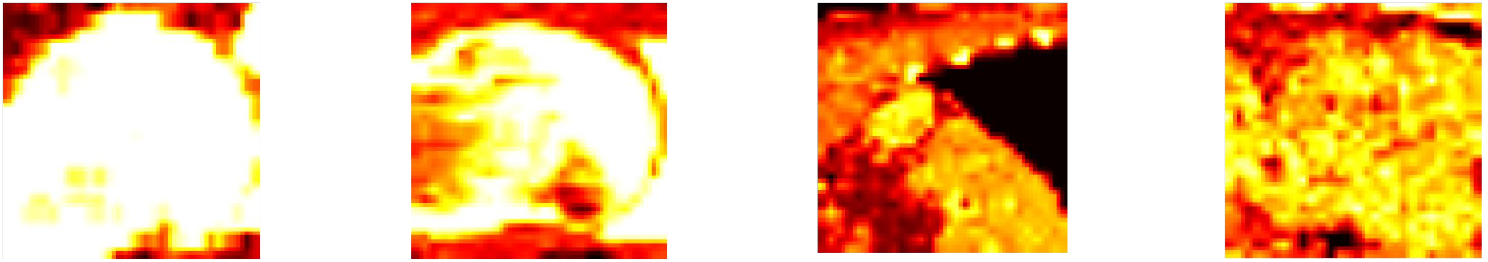} \\
        \midrule
        \makecell{OrganMNIST (3D)} & 11  & 1,742 & 10 & \includegraphics[width=0.4\textwidth]{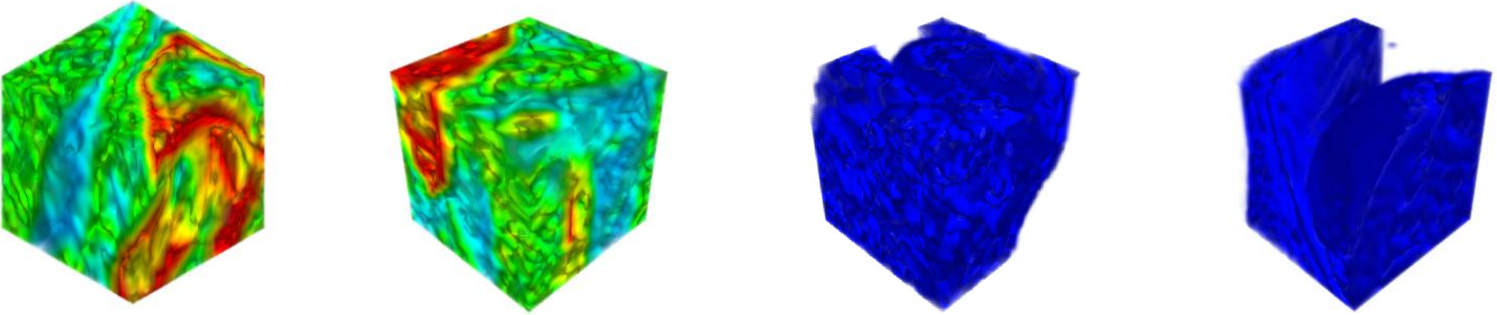} \\
        \bottomrule
    \end{tabular}}
    \label{tab:dataset_exm}
\end{table}

\begin{figure*}[tbh]
    \centering
    \includegraphics[width=0.85\textwidth]{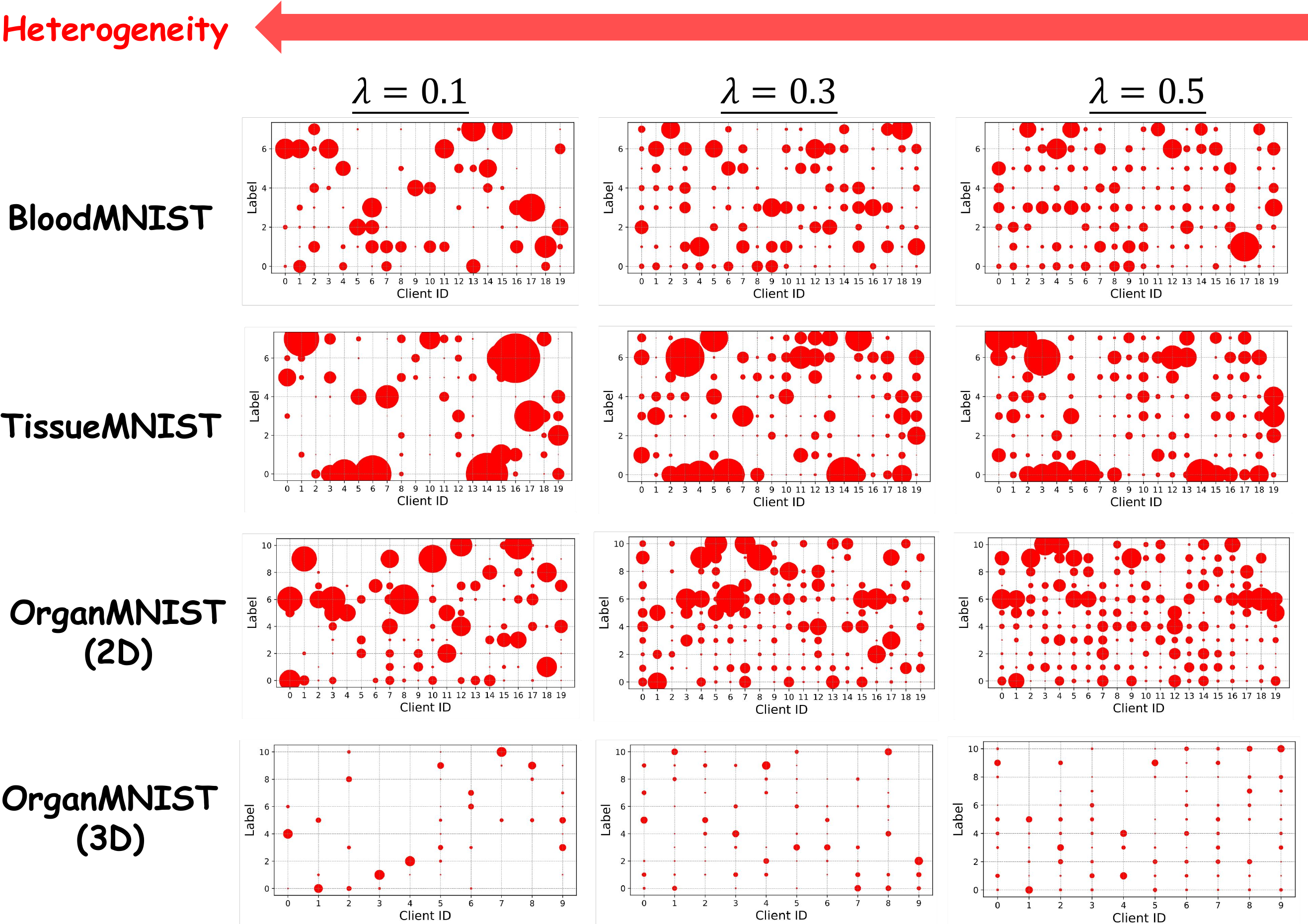}
    \caption{Visualization of four datasets distribution in different degrees of Non-IID environments. From top to bottom are BloodMNIST, TissueMNIST, OrganMNIST (2D) and OrganMNIST (2D), and from left to right are the Dirichlet distribution parameters $\lambda \in \{0.1, 0.3, 0.5\}$. The smaller $\lambda$ is, the stronger the data Non-IID between clients. In addition, a larger red circle means that the client has more such samples, and the opposite means that there are fewer such samples on the client.}
    \label{fig:dataset_distribution}
\end{figure*}

To demonstrate the effectiveness of our proposed \textsc{FedMIC} in real-world non-IID scenarios, we simulated a typical FL scenario by assigning each client's data according to a Dirichlet distribution \citep{wu2023bold}, specifically expressed as $\rvq \sim \texttt{Dirichlet}(\lambda \rvp)$. Here, $\rvp$ represents the prior class distribution, and $\lambda$ is a factor that modulates the degree of non-IID characteristics. A larger $\lambda$ value signifies more pronounced class imbalances within each client, leading to more demanding local tasks characterized by a greater variety of classes and fewer samples per class. As $\lambda$ increases, inter-client data distribution discrepancies decrease, while intra-client data distribution diversity increases.  This setup rigorously tests the robustness of methods against complex non-IID conditions. We conducted experiments using three Dirichlet parameters, $\lambda \in \{0.1, 0.3, 0.5\}$, to assess the impact of varying degrees of non-IID scenarios. The distribution of client data at different degrees of Non-IID is shown in Figure~\ref{fig:dataset_distribution}.

\subsection{Baselines}
We have selected the well-known SOTA FL algorithms, including a personalized algorithm based on parameter localized (\textsc{FedBN}), based on local regularization (\textsc{PerFedAvg}), based on knowledge distillation (\textsc{FedBE}), and \textsc{FedALA} as the baseline, to demonstrate the superiority of our \textsc{FedMIC} in MIC classification task. Detailed information of these baselines is as follows:

\textbf{\textsc{Local.}} In the absence of a distributed learning environment, medical images are processed independently by individual clients, precluding any information exchange between them.

\textbf{\textsc{FedAvg.}}~\citep{mcmahan2017communication} Aggregating local models to obtain a global model via classic average strategy while preserving the privacy of each individual's data.

\textbf{\textsc{FedBN.}}~\citep{li2021fedbn} A personalized FL approach that personalized batch-norm layers in local models and shares remaining parameters globally to achieve highly customized models for each client.

\textbf{\textsc{PerFedAvg.}}~\citep{t2020personalized} A personalized FL approach that adapts the global model to each user’s local data distribution while taking into account the similarity between users to improve model generalization.

\textbf{\textsc{FedBE.}}~\citep{chen2021fedbe} A personalized FL approach using Bayesian model integration for robustness in aggregating user predictions and summarizing integrated predictions into global models with the help of knowledge distillation.

\textbf{\textsc{FedALA.}}~\citep{zhang2023fedala} An approach for personalized federated learning that addresses inter-client statistical heterogeneity through its core component, the Adaptive Local Aggregation (ALA) module.

\textbf{\textsc{FedMIC.}} Our proposed framework introduces Dual Knowledge Distillation (Dual-KD) for local updating and Global Parameter Decomposition (GPD) to achieve a robust global model while maintaining highly customized local models for individual clients under privacy-preserving conditions.

To ensure a fair comparison, we maintained uniform baseline parameters as reported in the original publications. For the local model on each client, we selected \textbf{ResNet101}~\citep{he2016deep}. During training, we configured 50 global communication rounds and 5 local update epochs. We employed SGD as the optimizer, setting the local learning rate to $1 \times 10^{-3}$ for both teacher and student models. The performance of the framework was evaluated based on accuracy (\textbf{ACC}).

\subsection{Main Experiments and Results}
In this subsection, we will presents our main experiments on Non-IID environments with different degrees, including $\lambda = 0.1$ (\textbf{Table.~\ref{tab:lam1}}), $\lambda = 0.3$ (\textbf{Table.~\ref{tab:lam3}}), and $\lambda = 0.5$ (\textbf{Table.~\ref{tab:lam5}}). We will analyse the results for three different heterogeneous scenarios with different datasets as the basic unit.

\paragraph{Main Results on $\lambda = 0.1$.} The main results is shown in \textbf{Table.~\ref{tab:lam1}}, which demonstrate that our proposed \textsc{FedMIC} achieves the best performance across all four datasets and participation rate settings, significantly outperforming the baselines. \textbf{(1) BloodMNIST:} \textsc{FedMIC} achieves 92.13\%, 94.84\% and 95.11\% accuracy at 10\%, 30\% and 50\% participation rates, respectively. Compared to the second best method (FedALA at 10\% participation rate, 81.79\%), \textsc{FedMIC} achieved a relative performance improvement of 12.6\%. At 50\% participation rate, \textsc{FedMIC} improved accuracy by 4.2\% relative to FedALA (91.26\%). \textbf{(2) TissueMNIST:} \textsc{FedMIC} achieves accuracies of 64.28\%, 69.90\%, and 75.83\% at different participation rates. At a 10\% participation rate, \textsc{FedMIC} improves performance by 23.3\% compared to the second-best method, FedBE (52.12\%). At a 50\% participation rate, \textsc{FedMIC} improves accuracy by 10.1\% compared to FedALA (68.88\%). \textbf{(3) OrganMNIST (2D):} \textsc{FedMIC} achieves 85.55\% accuracy at a 50\% participation rate, an 8.5\% improvement over the second-best method (Local, 78.82\%). At a 10\% participation rate, \textsc{FedMIC} (70.21\%) achieves a 7.2\% improvement relative to FedALA (65.47\%). \textbf{(4) OrganMNIST (3D):} The advantages of \textsc{FedMIC} are even more pronounced. It achieves an accuracy of 94.62\% at a 50\% participation rate, a 6.6\% improvement relative to the second-best method (Local, 88.79\%). At a 10\% participation rate, \textsc{FedMIC} (76.62\%) improves accuracy by 15\% over FedALA (66.64\%).
\input{Table/dir1_res}

\paragraph{Main Results on $\lambda = 0.3$.} The main results is shown in \textbf{Table.~\ref{tab:lam3}}. \textbf{(1) BloodMNIST:} \textsc{FedMIC} performed exceptionally well, achieving the best performance at all participation rates. Notably, at a low participation rate of 10\%, \textsc{FedMIC} achieved an accuracy of 92.56\%, an improvement of 27.5\% compared to the second-best method (Local, 72.57\%), demonstrating its significant advantage in low participation rate scenarios. Compared to the case of $\lambda=0.1$, \textsc{FedMIC}'s performance on this dataset slightly improves, particularly at a 30\% participation rate, where the accuracy increases from 94.84\% to 96.84\%. \textbf{(2) TissueMNIST:} \textsc{FedMIC} also maintained its leadership position, achieving accuracies of 66.12\%, 70.37\%, and 77.27\% at various participation rates. Although the advantage over other methods is less pronounced than on BloodMNIST, \textsc{FedMIC} still performs best at all participation rates. Compared to $\lambda=0.1$, \textsc{FedMIC} shows a small improvement in performance on this dataset, specifically from 75.83\% to 77.27\% at 50\% participation rate. \textbf{(3) OrganMNIST (2D):} The performance of \textsc{FedMIC} fluctuated slightly. While it still performs best at 10\% and 30\% participation, achieving 73.64\% and 80.12 per cent accuracy, respectively, the Local method (82.99\%) slightly outperforms \textsc{FedMIC} (86.96\%) at 50\% participation. \textbf{(4)} \textsc{FedMIC} again maintains the best performance at all participation rates, achieving 79.21\%, 89.00\%, and 95.85\% accuracy, respectively. Compared to $\lambda = 0.1$, \textsc{FedMIC}'s performance on this dataset improves significantly, especially from 76.62\% to 79.21\% at a 10\% participation rate. At a 10\% participation rate, \textsc{FedMIC} improves its performance by 9.6\% relative to the second-best method, Local (72.26\%), further proving its superiority in low participation rate scenarios.
\input{Table/dir3_res}

\paragraph{Main Results on $\lambda=0.5$.} The results is shown in \textbf{Table.~\ref{tab:lam5}}. \textbf{(1) BloodMNIST:} \textsc{FedMIC} again maintains the best performance at all participation rates, achieving 79.21\%, 89.00\%, and 95.85\% accuracy, respectively. Compared to $\lambda = 0.1$, \textsc{FedMIC}'s performance on this dataset improves significantly, especially from 76.62\% to 79.21\% at 10\% participation rate. At 10\% participation rate, \textsc{FedMIC} improves its performance by 9.6\% relative to the second best method, Local (72.26\%), again proving its superiority in low participation rate scenarios. \textbf{(2) TissueMNIST:} \textsc{FedMIC} performed best at 30\% and 50\% participation rates, achieving 73.12\% and 79.12\% accuracy, respectively. At 10\% participation rate, \textsc{FedMIC} remains competitive although its accuracy (67.18\%) is slightly lower than FedALA (69.67\%). \textbf{(3) OrganMNIST (2D):} \textsc{FedMIC} performed best at 10\% and 30\% participation rates, achieving 74.49\% and 83.11\% accuracy, respectively. At 50\% participation rate, although \textsc{FedMIC} (89.24\%) is slightly lower than the Local method (88.25\%), it still outperforms the other methods. This result is similar to the case of $\lambda = 0.3$, suggesting that a high participation rate may make simple local training more effective under some specific data distributions. \textbf{(4) OrganMNIST (3D):} \textsc{FedMIC} again maintains the best performance at all participation rates, achieving 81.04\%, 89.48\%, and 96.33\% accuracy, respectively. Compared to $\lambda = 0.3$, \textsc{FedMIC}'s performance on this dataset shows further improvement, especially at high participation rates. At 10\% participation rate, \textsc{FedMIC} improves its performance by 4.4\% relative to the second best method, Local (77.65\%), again proving its superiority in low participation rate scenarios.
\input{Table/dir5_res}

\paragraph{General Analysis and Discussion.} Based on the above quantitative analysis for the results at different $\lambda$, we can find that: (1) In most cases, \textsc{FedMIC} demonstrates superior performance to other benchmark methods. This advantage is demonstrated for different datasets, participation rates and $\lambda$ values, proving the robustness and adaptability of the \textsc{FedMIC}. In particular, \textsc{FedMIC} consistently leads in scenarios dealing with low participation rates, showing its superior performance in federated learning environments with limited client participation. (2) For all values of $\lambda$, the performance of \textsc{FedMIC} generally shows an upward trend as the participation rate increases (from 10\% to 50\%). This suggests that \textsc{FedMIC} is able to effectively utilise more client data to improve model performance, while also maintaining its advantages at low participation rates. (3) Despite fluctuating performance under different conditions, \textsc{FedMIC} generally showed good stability and consistency.  In conclusion, the superiority of \textsc{FedMIC} under different medical image classification datasets comes from the dual knowledge distillation mechanism and global parameter decomposition mechanism in which we propose. The former provides powerful global knowledge to the local student model while also providing it with personalised insights into the local data, and each client storing heterogeneous medical image data is provided with an independent and highly personalised model, which is superior to training a separate model for each client because our framework breaks down the data silos and facilitates knowledge fusion. In addition, the latter provides a lightweight filtering strategy for the global aggregation process, which avoids bias in global knowledge by decomposing client parameters and retaining as much valid information as possible. These two mechanisms complement each other and further enhance the performance of our \textsc{FedMIC} in different tasks and scenarios.

\subsection{Hyperparameter Sensitivity}

We examined the impact of parameter $\alpha$ on \textsc{FedMIC} performance using the Global Parameter Decomposition strategy. Three additional values ($\alpha \in \{0.90, 0.92, 0.95\}$) were tested alongside the standard value of $\alpha = 0.98$ across four datasets. \textbf{Table~\ref{tab:paramstudy}} presents the results, which indicate that $\alpha = 0.98$ yields optimal performance. Lower $\alpha$ values resulted in reduced effectiveness due to insufficient parameter exchange between client and server, leading to loss of critical information. Conversely, higher values led to excessive exchange, introducing superfluous information.

\input{Table/param_impact}

\subsection{Ablation Studies}
We demonstrate the effectiveness and the necessity of the important components of our method through a series of ablation studies. Our \textsc{FedMIC}'s key innovations are in knowledge distillation and parameter decomposition, so we conduct ablation studies from two perspectives: (1) KD-based Ablation and (2) Parameter Decomposition-based Ablation. Here's the detialed setup for them:
\paragraph{KD-based Ablation} We evaluate the impact of our dual knowledge distillation mechanism by conducting experiments without certain parts:
\begin{itemize}
    \item \textsc{FedMIC}-A: This version omits the Distillation Auxiliary Matrix for local models. The correction loss is simply the MSE: $\mathcal{L}_{rep} = \frac{1}{n} \sum_{i=1}^{n}(\mathbf{H}^s_{hs} - \mathbf{H}^t_{hs})$.
    \item \textsc{FedMIC}-B: This version removes the representation-based DDL. The local loss is defined as $\mathcal{L}^t = \mathcal{L}^t_{dec-d} + \mathcal{L}^t_{task}$ and $\mathcal{L}^s = \mathcal{L}^s_{dec-d} + \mathcal{L}^s_{task}$ for the local teacher and student models, respectively.  \textit{Why representation-based DDL instead of decision-based DDL?} \textit{The latter encompasses the framework's critical distillation term, and its removal would substantially compromise the integrity of key procedural steps.}
\end{itemize}

\paragraph{Parameter Decomposition-based Ablation} We explore the effectiveness of our parameter decomposition strategy: 
\begin{itemize}
    \item \textsc{FedMIC}-C: This scenario removes Global Parameters Decomposition. As a result, full client model parameters are transmitted between clients and the server instead of just a subset of lightweight local model parameters.
\end{itemize}

\input{Table/abla_dir1_res}
\input{Table/abla_dir3_res}
\input{Table/abla_dir5_res}

We conducted ablation studies for three scenarios, consistent with previous experiments, to assess \textsc{FedMIC} at 10\%participation rate. These results are presented in Table~\ref{tab:ablation_dir01} (for $\lambda=0.1$), Table~\ref{tab:ablation_dir03} (for $\lambda=0.3$), and Table~\ref{tab:ablation_dir05} and (for $\lambda=0.5$). From these tables, we observed that: \textbf{(1)} In the KD-based ablation, standard \textsc{FedMIC} beats \textsc{FedMIC}-A and \textsc{FedMIC}-B. This indicates a drop in performance \textit{without the Auxiliary Matrix and Representatioon-based DDL}, highlighting their importance in our \textsc{FedMIC}; \textbf{(2)} For the parameter decomposition-based ablation, \textsc{FedMIC}-C, which sends full model parameters, increased the data exchange by about 88.51\%, leading to more communication overhead and reduced efficiency. The performance gains were marginal or even negative, showing that transmitting an additional \textbf{88.51\%} of parameters is not cost-effective; In summary, these ablation studies confirm the significance and effectiveness of each component in our \textsc{FedMIC}.

\subsection{Differential Privacy}
To further safeguard healthcare organization privacy in the proposed \textsc{FedMIC}, we implement differential privacy (DP) ~\citep{dwork2006differential} by adding random Gaussian noise to the model's gradients during global aggregation, as suggested by ~\citep{chen2023spatial}. We evaluated the performance of \textsc{FedMIC} both with and without DP. The noise, scaled by a factor $\tau$, is added to the shared parameters. We tested $\tau$ values of $\{1e^{-3}, 1e^{-2}, 5e^{-2}\}$ to apply differential privacy at varying levels in \textsc{FedMIC} and compared it with the standard FedAvg.
\input{Table/dp_res}
Table~\ref{tab:dpexp} presents the performance of \textit{FedMIC} and baseline methods on four fine-grained classification datasets under the Non-IID setting with $\lambda=0.1$ and a client participation rate of 10\%. The results indicate a slight decrease in \textit{FedMIC}'s performance after implementing DP. Nevertheless, \textit{FedMIC} continues to outperform standard FedAvg across all four datasets. As \textit{FedMIC} only transmits a subset of parameters during client-server communication, adding noise to these parameters suffices to maintain privacy security. Moreover, the performance degradation due to DP is less pronounced for FedMIC compared to baseline methods that exchange full model parameters during client-server communication.

\paragraph{Additional Discussion about Privacy} FL can be vulnerable to data breaches even when data is not directly shared among clients during global model training. An attacker might infer the original data from client-sent gradients, especially with small batch sizes and local training steps. Our \textsc{FedMIC} approach mitigates this risk by employing a local model for each participant, comprising teacher and student models for private data training. We implement a cost-effective parameter decomposition strategy during client-server communication, transmitting only a subset of parameters. This approach impedes attackers' ability to reconstruct original data from partial gradients. Furthermore, we enhance the strategy with adjustable control coefficients $\alpha$, further complicating potential inference attacks on raw data, even as the number of training rounds approaches infinity: $e \rightarrow \infty$.

\section{Conclusion and Future Works}
This paper introduces a novel privacy-preserving framework for medical image classification (\textsc{FedMIC}), combining dual knowledge distillation (\textbf{Dual-KD}) with global parameter decomposition (\textbf{GPD}). The Dual-KD enables clients to leverage a rich global knowledge base, enhancing personalized representation of local data and offering a robust, customized solution for varied environments. The GPD strategy reduces communication costs by requiring participants to upload only select parameters rather than entire models, particularly beneficial in resource-constrained settings. Extensive evaluation on real-world medical image classification (including 2D and 3D) datasets confirms the effectiveness and advantages of our proposed \textsc{FedMIC}.

\paragraph{Limitations} Our work has two main limitations: (1) Training both student and teacher models on each client can strain local resources, especially with large models, and (2) The effectiveness of the low-rank decomposition process in reducing upload size may diminish when student model parameter dimensions across clients are not widely separated or consistent.

\paragraph{Future Work} Future research will focus on two areas: (1) \textbf{Framework level:} We aim to further reduce client-server communication costs while maintaining high performance in real-world medical imaging applications and (2) \textbf{Application level:} We plan to extend the framework to a wider range of applications, particularly in visual language tasks, including multimodal medical data retrieval and fusion. Our goal is to develop cost-effective solutions for AI-related medical applications in resource-constrained environments.

\small
\bibliographystyle{unsrtnat}
\bibliography{reference}


\end{document}

%% file: math_commands.tex
\usepackage{amsmath,amsfonts,bm}

\def\eqref#1{equation~\ref{#1}}

\def\1{\bm{1}}

\def\rvg{{\mathbf{g}}}

\def\rvp{{\mathbf{p}}}
\def\rvq{{\mathbf{q}}}
\def\rvr{{\mathbf{r}}}

\def\mH{{\bm{H}}}

\def\mP{{\bm{P}}}

\def\mW{{\bm{W}}}
\def\mX{{\bm{X}}}
\def\mY{{\bm{Y}}}

\DeclareMathAlphabet{\mathsfit}{\encodingdefault}{\sfdefault}{m}{sl}
\SetMathAlphabet{\mathsfit}{bold}{\encodingdefault}{\sfdefault}{bx}{n}

\def\gD{{\mathcal{D}}}

\def\gH{{\mathcal{H}}}

\def\gL{{\mathcal{L}}}

\def\sP{{\mathbb{P}}}

\def\sR{{\mathbb{R}}}

\def\emU{{U}}
\def\emV{{V}}

\DeclareMathOperator*{\argmin}{arg\,min}

%% file: Table/dir1_res.tex
\begin{table}[tbh]
  \centering
    \caption{Experimental results of the proposed \textsc{FedMIC} and baseline for four MIC datasets with $\lambda=0.1$ and client participant ratio $\rvr \in \{10\%, 30\%, 50\%\}$, where \textbf{Bold} are the best, and \underline{underline} represent the second best.}
  \resizebox{\textwidth}{!}{
    \begin{tabular}{c|ccc|ccc|ccc|ccc}
    \toprule
    \textsc{Dataset} & \multicolumn{3}{c|}{BloodMNIST} & \multicolumn{3}{c|}{TissueMNIST} & \multicolumn{3}{c|}{OrganMNIST (2D)} & \multicolumn{3}{c}{OrganMNIST (3D)} \\
    \midrule
    Method/Ratio & 10\% & 30\% & 50\% & 10\% & 30\% & 50\% & 10\% & 30\% & 50\% & 10\% & 30\% & 50\% \\
    \midrule
    \multirow{2}[0]{*}{\textsc{FedAvg}} & 49.92  & 58.12  & 60.23  & 44.37  & 52.55  & 58.75  & 59.82  & 64.05  & 69.92  & 55.17  & 61.92  & 62.35  \\
    & \small $\pm 1.86$ & \small $\pm 0.57$ & \small $\pm 1.71$ & $\pm 1.24$ & \small $\pm 1.95$ & \small $\pm 1.12$ & \small $\pm 0.33$ & \small $\pm 0.75$ & \small $\pm 0.17$ & \small $\pm 1.43$ & \small $\pm 0.04$ & \small $\pm 1.58$ \\
    \multirow{2}[0]{*}{\textsc{Local}} & 68.35 & 88.32 & \underline{92.85} & 57.49 & 59.85 & 66.61 & \underline{69.23} & \underline{76.42} & \underline{78.82} & 68.47 & \underline{86.18} & \underline{88.79} \\
       & \small $\pm1.93$ & \small $\pm1.23$ & \small $\pm0.39$ & \small $\pm2.76$ & \small $\pm0.64$ & \small $\pm1.05$ & \small $\pm1.37$ & \small $\pm0.84$ & \small $\pm2.14$ & \small $\pm1.46$ & \small $\pm1.61$ & \small $\pm2.05$ \\
    \multirow{2}[0]{*}{\textsc{FedBN}} & 79.28 & 82.45 & 87.69 & 58.55 & \underline{64.35} & 67.82 & 66.12 & 71.49 & 75.37 & 58.24 & 72.64 & 76.18 \\
       & \small $\pm1.47$ & \small $\pm2.51$ & \small $\pm0.09$ & \small $\pm0.81$ & \small $\pm2.26$ & \small $\pm0.45$ & \small $\pm1.11$ & \small $\pm1.14$ & \small $\pm1.87$ & \small $\pm2.64$ & \small $\pm0.87$ & \small $\pm2.32$ \\
    \multirow{2}[0]{*}{\textsc{PerFedAvg}} & 77.81 & 87.32 & 90.55 & 54.22 & 60.88 & 64.85 & 63.47 & 72.15 & 74.28 & 62.19 & 78.74 & 81.12 \\
       & \small $\pm1.19$ & \small $\pm0.93$ & \small $\pm0.64$ & \small $\pm1.68$ & \small $\pm2.75$ & \small $\pm0.51$ & \small $\pm0.82$ & \small $\pm0.23$ & \small $\pm0.54$ & \small $\pm1.96$ & \small $\pm1.10$ & \small $\pm0.37$ \\
    \multirow{2}[0]{*}{\textsc{FedBE}} & 81.24  &  88.21  & 89.07  & 52.12  & 63.43 &  67.90 & 63.64  &  69.52  &  74.22  & 64.41  &  78.55 & 84.20  \\
       & \small $\pm1.90$ & \small $\pm0.88$ & \small $\pm1.46$ & \small $\pm1.13$ & \small $\pm0.77$ & \small $\pm2.10$ & \small $\pm1.17$ & \small $\pm1.27$ & \small $\pm1.21$ & \small $\pm1.37$ & \small $\pm1.26$ & \small $\pm1.21$ \\
    \multirow{2}[0]{*}{\textsc{FedALA}} & \underline{81.79}  & \underline{89.47}  & 91.26  & \underline{59.11}  & 64.00  & \underline{68.88}  & 65.47  & 70.15  &  76.02  & \underline{66.64}  & 76.31  & 83.21  \\
       & \small $\pm0.42$ & \small $\pm0.61$ & \small $\pm0.54$ & \small $\pm0.58$ & \small $\pm0.04$ & \small $\pm0.17$ & \small $\pm0.62$ & \small $\pm1.47$ & \small $\pm0.66$ & \small $\pm1.07$ & \small $\pm0.44$ & \small $\pm1.62$ \\
    \midrule
    \multirow{2}[0]{*}{\textbf{\textsc{FedMIC}  (Ours)}} & \bf 92.13  & \bf 94.84  & \bf 95.11  & \bf 64.28  & \bf 69.90  & \bf 75.83  & \bf 70.21  & \bf 77.42  & \bf 85.55  & \bf 76.62  & \bf 87.18  & \bf 94.62  \\
    & \small $\pm 1.46$ & \small $\pm 0.27$ & \small $\pm 1.73$ & \small $\pm 0.78$ & \small $\pm 0.42$ & \small $\pm 1.12$ & \small $\pm 0.62$ & \small $\pm 0.19$ & \small $\pm 0.98$ & \small $\pm 1.46$ & \small $\pm 0.46$ & \small $\pm 0.99$ \\
    \bottomrule
    \end{tabular}}
  \label{tab:lam1}
\end{table}

%% file: Table/dir3_res.tex
\begin{table}[tbh]
  \centering
    \caption{Experimental results of the proposed PRFL and baseline for four MIC datasets with $\lambda=0.3$ and client participant ratio $\rvr \in \{10\%, 30\%, 50\%\}$, where \textbf{Bold} are the optimal results, and \underline{underline} represent suboptimal result.}
  \resizebox{\textwidth}{!}{
    \begin{tabular}{c|ccc|ccc|ccc|ccc}
    \toprule
    \textsc{Dataset} & \multicolumn{3}{c|}{BloodMNIST} & \multicolumn{3}{c|}{TissueMNIST} & \multicolumn{3}{c|}{OrganMNIST (2D)} & \multicolumn{3}{c}{OrganMNIST (3D)} \\
    \midrule
    Method/Ratio & 10\% & 30\% & 50\% & 10\% & 30\% & 50\% & 10\% & 30\% & 50\% & 10\% & 30\% & 50\% \\
    \midrule
    \multirow{2}[0]{*}{\textsc{FedAvg}} & 51.85  & 60.90  & 63.09  & 46.48  & 56.64  & 62.22  & 61.93  & 68.56  & 71.91  & 58.76  & 65.36  & 66.28  \\
       & \small $\pm1.20$ & \small $\pm0.66$ & \small $\pm1.79$ & \small $\pm0.52$ & \small $\pm1.33$ & \small $\pm0.85$ & \small $\pm1.09$ & \small $\pm0.97$ & \small $\pm0.31$ & \small $\pm1.50$ & \small $\pm0.44$ & \small $\pm0.76$ \\
    \multirow{2}[0]{*}{\textsc{Local}} & 72.57 & \underline{94.90} & \underline{95.78} & 60.63 & 67.16 & 71.50 & \textbf{75.57} & \bf 84.99 & 82.99 & \underline{72.26} & \bf 93.39 & \underline{94.35} \\
       & \small $\pm1.24$ & \small $\pm0.95$ & \small $\pm1.52$ & \small $\pm0.25$ & \small $\pm1.03$ & \small $\pm0.89$ & \small $\pm1.76$ & \small $\pm0.42$ & \small $\pm1.68$ & \small $\pm0.57$ & \small $\pm1.00$ & \small $\pm0.83$ \\
    \multirow{2}[0]{*}{\textsc{FedBN}} & 82.46 & 87.26 & 93.29 & 62.23 & \underline{70.43} & 73.66 & 70.45 & 77.31 & 82.00 & 60.92 & 80.68 & 84.50 \\
       & \small $\pm1.02$ & \small $\pm0.23$ & \small $\pm1.39$ & \small $\pm0.17$ & \small $\pm1.90$ & \small $\pm0.56$ & \small $\pm1.07$ & \small $\pm1.68$ & \small $\pm0.82$ & \small $\pm1.57$ & \small $\pm0.73$ & \small $\pm0.48$ \\
    \multirow{2}[0]{*}{\textsc{PerFedAvg}} & 81.52 & 90.94 & 91.62 & 58.90 & 65.37 & 70.48 & 69.35 & \underline{81.08} & \underline{83.33} & 70.39 & 87.30 & 87.74 \\
       & \small $\pm0.12$ & \small $\pm1.66$ & \small $\pm0.77$ & \small $\pm1.83$ & \small $\pm1.21$ & \small $\pm0.34$ & \small $\pm1.45$ & \small $\pm0.89$ & \small $\pm0.58$ & \small $\pm1.11$ & \small $\pm0.99$ & \small $\pm1.74$ \\
    \multirow{2}[0]{*}{\textsc{FedBE}} & \underline{83.25} & 89.62 & 90.88 & 62.52 & 65.58 & \underline{74.90} & 67.21 & 73.80 & 78.42 & 67.91 & 79.19 & 85.23 \\
       & \small $\pm0.56$ & \small $\pm1.37$ & \small $\pm0.85$ & \small $\pm0.33$ & \small $\pm1.94$ & \small $\pm0.20$ & \small $\pm1.26$ & \small $\pm1.50$ & \small $\pm0.65$ & \small $\pm1.88$ & \small $\pm0.91$ & \small $\pm0.44$ \\
    \multirow{2}[0]{*}{\textsc{FedALA}} & 82.95 & 89.63 & 92.46 & \underline{64.03} & 67.29 & 70.69 & \underline{74.07} & 74.62 & 79.90 & 70.69 & 79.02 & 85.13 \\
       & \small $\pm1.57$ & \small $\pm0.24$ & \small $\pm1.32$ & \small $\pm0.02$ & \small $\pm0.75$ & \small $\pm1.98$ & \small $\pm0.37$ & \small $\pm1.11$ & \small $\pm0.90$ & \small $\pm0.65$ & \small $\pm1.79$ & \small $\pm0.14$ \\
       \midrule
    \multirow{2}[0]{*}{\textbf{\textsc{FedMIC} (Ours)}} & \bf 92.56 & \bf 96.84 & \bf 96.41 & \bf 66.12 & \bf 70.37 & \bf 77.27 & 73.64 & 80.12 & \bf 86.96 & \bf 79.21 & \underline{89.00} & \bf 95.85 \\
       & \small $\pm1.27$ & \small $\pm0.33$ & \small $\pm1.90$ & \small $\pm0.57$ & \small $\pm1.64$ & \small $\pm0.68$ & \small $\pm1.02$ & \small $\pm0.45$ & \small $\pm1.78$ & \small $\pm1.05$ & \small $\pm0.12$ & \small $\pm0.96$ \\
    \bottomrule
    \end{tabular}}
  \label{tab:lam3}
\end{table}

%% file: Table/dir5_res.tex
\begin{table}[tbh]
  \centering
    \caption{Experimental results of our \textsc{FedMIC} and baseline for four MIC datasets with $\lambda=0.5$ and client participant ratio $\rvr \in \{10\%, 30\%, 50\%\}$, where \textbf{Bold} are the optimal results, and \underline{underline} represent suboptimal result.}
  \resizebox{\textwidth}{!}{
    \begin{tabular}{c|ccc|ccc|ccc|ccc}
    \toprule
    \textsc{Dataset} & \multicolumn{3}{c|}{BloodMNIST} & \multicolumn{3}{c|}{TissueMNIST} & \multicolumn{3}{c|}{OrganMNIST (2D)} & \multicolumn{3}{c}{OrganMNIST (3D)} \\
    \midrule
    Method/Ratio & 10\% & 30\% & 50\% & 10\% & 30\% & 50\% & 10\% & 30\% & 50\% & 10\% & 30\% & 50\% \\
    \midrule
    \multirow{2}[0]{*}{\textsc{FedAvg}} & 56.45  & 64.48  & 68.68  & 47.26  & 60.53  & 63.32  & 64.21  & 74.99  & 76.23  & 59.08  & 68.41  & \underline{71.23}  \\
       & \small $\pm1.34$ & \small $\pm0.87$ & \small $\pm1.12$ & \small $\pm0.65$ & \small $\pm1.56$ & \small $\pm1.43$ & \small $\pm0.98$ & \small $\pm1.75$ & \small $\pm0.29$ & \small $\pm1.68$ & \small $\pm0.52$ & \small $\pm1.21$ \\
    \multirow{2}[0]{*}{\textsc{Local}} & 77.92  & \underline{96.18}  & \underline{96.37}  & 65.69  & \bf 75.30  & \underline{76.48}  & \bf 77.72  & \bf 86.28  & \underline{88.25}  & \underline{77.65}  & \bf 94.56  & \underline{95.21}  \\
       & \small $\pm1.03$ & \small $\pm1.68$ & \small $\pm0.86$ & \small $\pm1.01$ & \small $\pm1.29$ & \small $\pm1.74$ & \small $\pm0.58$ & \small $\pm1.20$ & \small $\pm1.02$ & \small $\pm0.77$ & \small $\pm1.31$ & \small $\pm1.91$ \\
    \multirow{2}[0]{*}{\textsc{FedBN}} & 83.51  & 89.52  & 93.01  & 67.13  & 68.06  & 73.68  & 73.23  & 76.28  & 82.22  & 71.92  & 76.27  & 83.93  \\
       & \small $\pm1.94$ & \small $\pm1.23$ & \small $\pm1.41$ & \small $\pm0.06$ & \small $\pm1.11$ & \small $\pm0.57$ & \small $\pm1.12$ & \small $\pm1.76$ & \small $\pm0.98$ & \small $\pm1.81$ & \small $\pm1.27$ & \small $\pm1.54$ \\
    \multirow{2}[0]{*}{\textsc{PerFedAvg}} & 82.30  & 91.36  & 91.55  & 61.31  & \underline{68.23}  & 72.68  & 71.95  &  76.21  &  81.77  & 70.01  & 80.55  & 88.32  \\
       & \small $\pm1.12$ & \small $\pm0.68$ & \small $\pm1.57$ & \small $\pm0.99$ & \small $\pm1.40$ & \small $\pm1.15$ & \small $\pm0.86$ & \small $\pm1.34$ & \small $\pm0.72$ & \small $\pm1.08$ & \small $\pm0.53$ & \small $\pm1.66$ \\
    \multirow{2}[0]{*}{\textsc{FedBE}} & \underline{84.12}  & 90.10  & 91.94  & \underline{68.94}  & 69.20  & 73.27  & 68.45  & 75.58   &  81.79  & 69.62  & 80.38  & 86.61  \\
       & \small $\pm1.01$ & \small $\pm1.87$ & \small $\pm0.56$ & \small $\pm1.18$ & \small $\pm0.33$ & \small $\pm1.90$ & \small $\pm0.82$ & \small $\pm1.60$ & \small $\pm0.44$ & \small $\pm1.35$ & \small $\pm0.68$ & \small $\pm1.75$ \\
    \multirow{2}[0]{*}{\textsc{FedALA}} & 84.20  & 90.00  & 92.99 & \textbf{69.67}  & 70.60  & 73.31  & \underline{75.33}  & 76.28  &  82.62  & 73.22  & 81.27  & 87.12  \\
    & \small $\pm1.45$ & \small $\pm0.88$ & \small $\pm1.12$ & \small $\pm0.64$ & \small $\pm1.29$ & \small $\pm0.73$ & \small $\pm1.01$ & \small $\pm0.92$ & \small $\pm1.54$ & \small $\pm0.46$ & \small $\pm1.67$ & \small $\pm0.53$ \\
       \midrule
    \multirow{2}[0]{*}{\textbf{\textsc{FedMIC} (Ours)}} & \bf 93.42  & \bf 96.99  & \bf 97.97  & 67.18  & \underline{73.12}  & \bf 79.12  & \bf 74.49  & \underline{83.11}  & \bf 89.24  & \bf 81.04  & \underline{89.48}  & \bf 96.33  \\
    & \small $\pm1.45$ & \small $\pm1.21$ & \small $\pm0.98$ & \small $\pm0.62$ & \small $\pm1.30$ & \small $\pm1.88$ & \small $\pm1.15$ & \small $\pm0.89$ & \small $\pm0.34$ & \small $\pm1.73$ & \small $\pm0.47$ & \small $\pm1.60$ \\
    \bottomrule
    \end{tabular}}
  \label{tab:lam5}
\end{table}

%% file: Table/param_impact.tex
\begin{table}[tbh]
  \centering
    \caption{Experimental results on the performance with different $\alpha \in \{0.90, 0.92, 0.95, 0.98\}$, where \textbf{Bold} denotes the best, \underline{Underline} denotes the second best. $\alpha=0.98$ was the original setting.}
  \resizebox{0.8\textwidth}{!}{
    \begin{tabular}{c|c|c|c|c}
    \toprule
    $\alpha$  & BloodMNIST & TissueMNIST & OrganMNIST (2D) & OrganMNIST (3D) \\
    \midrule
    0.98 & \bf 93.42 \small $\pm 1.45$  & \bf 67.18 \small $\pm 0.62$ & \bf 74.49 \small $\pm 1.15$ & \bf \underline{81.04} \small $\pm 1.73$ \\
    \midrule
    0.90 & 92.11 \small $\pm 0.46$  & \underline{66.91} \small $\pm 0.18$ & 73.59 \small $\pm 1.22$  & 80.16 \small $\pm 0.21$\\
    0.92 & \underline{92.49} \small $\pm 1.46$ & 67.04 \small $\pm 0.82$ & 73.26 \small $\pm 1.02$  & 80.00 \small $\pm 1.72$\\
    0.95 & 91.21 \small $\pm 1.23$  & 66.20 \small $\pm 0.12$ & \underline{74.04} \small $\pm 1.22$ & \bf 81.22 \small $\pm 0.40$\\
    \bottomrule
    \end{tabular}}
  \label{tab:paramstudy}%
\end{table}%

%% file: Table/abla_dir1_res.tex
\begin{table}[tbh]
  \centering
    \caption{Comparison of performance under $\lambda= 0.1$, where $\textbf{Bold}$ denotes the best performance of incense under KD-based ablation and $\textbf{\textcolor{red}{Bold}}$ denotes the best result under parameters decomposition-based ablation experiments.} 
  \resizebox{.9\textwidth}{!}{
    \begin{tabular}{c|c|c|c|c}
    \toprule
    Method / Dataset & BloodMNIST & TissueMNIST & OrganMNIST (2D) & OrganMNIST (3D) \\
    \midrule
    \multirow{3}[1]{*}{\textsc{FedMIC}} & \bf 92.13  & \bf 64.28  & \bf 70.21  & \bf 76.62  \\
       & \small $\pm 1.46$ & \small $\pm 0.78$ & \small $\pm 0.62$ & \small $\pm 1.46$ \\
       & \multicolumn{4}{c}{\cellcolor{red! 30} \bf $11.49\%$ of Complete Model Parameters Communication} \\
    \midrule
    \multirow{3}[0]{*}{\textsc{FedMIC}-A} & 90.44 & 61.03 & 66.92 & 73.10 \\
       & \small $\pm 0.12$ & \small $\pm 0.45$ & \small $\pm 0.46$ & \small $\pm 0.63$ \\
       & \multicolumn{4}{c}{\cellcolor{red! 30} \bf $11.49\%$ of Complete Model Parameters Communication} \\
    \midrule
    \multirow{3}[1]{*}{\textsc{FedMIC}-B} & 91.00 & 60.21 & 63.26 & 70.82 \\
       & \small $\pm 0.42$ & \small $\pm 0.12$ & \small $\pm 1.12$ & \small $\pm 0.41$ \\
       & \multicolumn{4}{c}{\cellcolor{red! 30} \bf $11.49\%$ of Complete Model Parameters Communication} \\
    \midrule
    \multirow{3}[1]{*}{\textsc{FedMIC}-C} & \textcolor{red}{\bf 92.33} & \textcolor{red}{\bf 65.29} & \textcolor{red}{\bf 69.75} & \textcolor{red}{\bf 76.76} \\
       & \small $\pm 0.42$ & \small $\pm 0.32$ & \small $\pm 0.19$ & \small $\pm 0.67$ \\
       & \multicolumn{4}{c}{\cellcolor{green! 30} \bf $100\%$ of Complete Model Parameters Communication} \\
    \bottomrule
    \end{tabular}}
  \label{tab:ablation_dir01}
\end{table}

%% file: Table/abla_dir3_res.tex
\begin{table}[tbh]
  \centering
    \caption{Comparison of performance under $\lambda = 0.3$, where $\textbf{Bold}$ denotes the best performance of incense under knowledge distillation-based ablation experiments and $\textbf{\textcolor{red}{Bold}}$ denotes the best result under parameters decomposition-based ablation experiments.} 
  \resizebox{.9\textwidth}{!}{
    \begin{tabular}{c|c|c|c|c}
    \toprule
    Method / Dataset & BloodMNIST & TissueMNIST & OrganMNIST (2D) & OrganMNIST (3D) \\
    \midrule
    \multirow{3}[1]{*}{\textsc{FedMIC}} & \bf 92.56  & \bf 66.12  & \bf 73.64  & \bf 79.21  \\
       & \small $\pm 1.27$ & \small $\pm 0.57$ & \small $\pm 1.02$ & \small $\pm 1.05$ \\
       & \multicolumn{4}{c}{\cellcolor{red! 30} \bf $11.49\%$ of Complete Model Parameters Communication} \\
    \midrule
    \multirow{3}[0]{*}{\textsc{FedMIC}-A} & 90.66 & 62.24 & 68.22 & 75.22 \\
       & \small $\pm 0.12$ & \small $\pm 0.68$ & \small $\pm 0.24$ & \small $\pm 0.62$ \\
       & \multicolumn{4}{c}{\cellcolor{red! 30} \bf $11.49\%$ of Complete Model Parameters Communication} \\
    \midrule
    \multirow{3}[1]{*}{\textsc{FedMIC}-B} & 91.70 & 62.67 & 69.05 & 72.45 \\
       & \small $\pm 0.43$ & \small $\pm 0.98$ & \small $\pm 0.32$ & \small $\pm 0.29$ \\
       & \multicolumn{4}{c}{\cellcolor{red! 30} \bf $11.49\%$ of Complete Model Parameters Communication} \\
    \midrule
    \multirow{3}[1]{*}{\textsc{FedMIC}-C} & \textcolor{red}{\bf 92.22} & \textcolor{red}{\bf 67.43} & \textcolor{red}{\bf 74.02} & \textcolor{red}{\bf 80.20} \\
       & \small $\pm 0.12$ & \small $\pm 0.87$ & \small $\pm 0.62$ & \small $\pm 1.04$ \\
       & \multicolumn{4}{c}{\cellcolor{green! 30} \bf $100\%$ of Complete Model Parameters Communication} \\
    \bottomrule
    \end{tabular}}
  \label{tab:ablation_dir03}%
\end{table}

%% file: Table/abla_dir5_res.tex
\begin{table}[tbh]
  \centering
    \caption{Comparison of performance under $\lambda = 0.5$, where $\textbf{Bold}$ denotes the best performance of incense under knowledge distillation-based ablation experiments and $\textbf{\textcolor{red}{Bold}}$ denotes the best result under parameters decomposition-based ablation experiments.} 
  \resizebox{.9\textwidth}{!}{
    \begin{tabular}{c|c|c|c|c}
    \toprule
    Method / Dataset & BloodMNIST & TissueMNIST & OrganMNIST (2D) & OrganMNIST (3D) \\
    \midrule
    \multirow{3}[1]{*}{\textsc{FedMIC}} & \bf 93.42  & \bf 67.18  & \bf 74.49  & \bf 87.04  \\
       & \small $\pm 0.43$ & \small $\pm 0.24$ & \small $\pm 0.65$ & \small $\pm 0.23$ \\
       & \multicolumn{4}{c}{\cellcolor{red! 30} \bf $11.49\%$ of Complete Model Parameters Communication} \\
    \midrule
    \multirow{3}[0]{*}{\textsc{FedMIC}-A} & 90.62 & 63.21 & 69.36 & 81.00 \\
       & \small $\pm 0.42$ & \small $\pm 0.16$ & \small $\pm 0.20$ & \small $\pm 1.12$ \\
       & \multicolumn{4}{c}{\cellcolor{red! 30} \bf $11.49\%$ of Complete Model Parameters Communication} \\
    \midrule
    \multirow{3}[1]{*}{\textsc{FedMIC}-B} & 91.23 & 63.34 & 69.89 & 79.33 \\
       & \small $\pm 0.21$ & \small $\pm 0.54$ & \small $\pm 0.42$ & \small $\pm 0.56$ \\
       & \multicolumn{4}{c}{\cellcolor{red! 30} \bf $11.49\%$ of Complete Model Parameters Communication} \\
    \midrule
    \multirow{3}[1]{*}{\textsc{FedMIC}-C} & \textcolor{red}{\bf 93.90} & \textcolor{red}{\bf 68.12} & \textcolor{red}{\bf 75.21} & \textcolor{red}{\bf 88.04} \\
       & \small $\pm 0.22$ & \small $\pm 0.89$ & \small $\pm 0.73$ & \small $\pm 0.48$ \\
       & \multicolumn{4}{c}{\cellcolor{green! 30} \bf $100\%$ of Complete Model Parameters Communication} \\
    \bottomrule
    \end{tabular}}
  \label{tab:ablation_dir05}%
\end{table}

%% file: Table/dp_res.tex
\begin{table}[H]
  \centering
    \caption{Results of Differential Privacy experiments (DP factor $\tau \in \{1e^{-3}, 1e^{-2}, 5e^{-2}\}$) of the proposed \textsc{FedMIC} and baseline methods under four fine-grained classification datasets under the Dirichlet Non-IID setting ($\lambda=0.1$), where \textbf{Bold} denotes the optimal result, \underline{Underline} denotes the suboptimal result, $\downarrow$ denotes the degradation of the performance, with lower being better. Note that \textit{w} and \textit{wo} present \textit{with} and \textit{without}, respectively.}
  \resizebox{\textwidth}{!}{
    \begin{tabular}{c|c|c|c|c|c}
    \toprule
    \sc Method & \sc Differential Privacy & BloodMNIST & TissueMNIST & OrganMNIST (2D) & OrganMNIST (3D) \\
    \midrule
    \multirow{5}[2]{*}{\sc FedAvg} & \textit{wo}  &  49.92$_{\pm 1.86}$  &  44.37$_{\pm 1.24}$  & 59.82$_{\pm 0.33}$  &  55.17$_{\pm 1.43}$\\
    & \textit{w} ($\tau=1e^{-3}$) &  46.83$_{\pm 0.75}$  &  41.95$_{\pm 0.62}$  &  56.59$_{\pm 0.42}$  & 52.44$_{\pm 1.40}$ \\
    & \textit{w} ($\tau=1e^{-2}$) &  43.78$_{\pm 0.42}$  &  39.52$_{\pm 0.62}$  &  54.31$_{\pm 0.19}$  & 50.16$_{\pm 0.29}$ \\
    & \textit{w} ($\tau=5e^{-2}$) &  39.96$_{\pm 0.56}$  &  34.83$_{\pm 0.63}$  &  51.24$_{\pm 0.11}$  & 45.62$_{\pm 0.31}$ \\
    & \bf Mean Variation &  6.40 $\downarrow$  &  5.60 $\downarrow$ & 5.77 $\downarrow$   &  5.76 $\downarrow$\\
    \midrule
    \multirow{5}[2]{*}{\sc \bf FedMIC (Ours)} &  \textit{wo} &  92.13$_{\pm 1.46}$  &  64.28$_{\pm 0.78}$  & 70.21$_{\pm 0.62}$  &  76.62$_{\pm 1.46}$ \\
       & \textit{w} ($\tau=1e^{-3}$) &  90.87$_{\pm 0.71}$  &  62.69$_{\pm 0.21}$  &  68.53$_{\pm 0.45}$  &  74.95$_{\pm 0.12}$ \\
       & \textit{w} ($\tau=1e^{-2}$) &  89.45$_{\pm 0.33}$  &  61.32$_{\pm 1.21}$  &  67.08$_{\pm 1.09}$  &  73.41$_{\pm 1.32}$ \\
       & \textit{w} ($\tau=5e^{-2}$) &  87.92$_{\pm 1.27}$  &  59.76$_{\pm 0.42}$  &  65.47$_{\pm 0.26}$  &  71.83$_{\pm 0.10}$ \\
        & \bf Mean Variation &  \bf 4.21 $\downarrow$  &  \bf 4.52 $\downarrow$ & \bf 4.74 $\downarrow$   &  \bf 4.79 $\downarrow$\\
    \bottomrule
    \end{tabular}}
  \label{tab:dpexp}
\end{table}%